\documentclass[11pt]{article}
\usepackage{graphicx}
\usepackage{xspace}
\usepackage[preprint]{acl}

\usepackage{times}
\usepackage{latexsym}
\usepackage{tabularx} 
\usepackage[T1]{fontenc}

\usepackage[utf8]{inputenc}

\usepackage{microtype}
\usepackage{verbatim}
\usepackage{inconsolata}

\usepackage{graphicx}
\usepackage{xspace}
\usepackage[table]{xcolor}
\definecolor{OpenAITeal}{HTML}{10A37F}
\definecolor{DeepSeekBlue}{HTML}{0047AB}
\definecolor{QwenIndigo}{HTML}{5C46E5}
\definecolor{NousOrange}{HTML}{EA580C}
\usepackage{colortbl} 
\usepackage{pgfplots}
\pgfplotsset{compat=1.18}
\definecolor{darkblue}{rgb}{0, 0, 0.5}
\hypersetup{colorlinks=true, citecolor=darkblue, linkcolor=darkblue, urlcolor=darkblue}
\usepackage{tikz}
\usepackage{amssymb}
\usepackage{amsmath}
\usepackage{subcaption}
\usepackage{booktabs}
\usepackage{algorithm}
\usepackage{algpseudocode}
\usepackage{listings}
\AtBeginDocument{%
  \setlength{\abovedisplayskip}{3pt}%
  \setlength{\belowdisplayskip}{3pt}%
  \setlength{\abovedisplayshortskip}{3pt}%
  \setlength{\belowdisplayshortskip}{3pt}%
}

\title{Courtroom-Style Multi-Agent Debate with Progressive RAG and Role-Switching for Controversial Claim Verification}

\author{
  Masnun Nuha Chowdhury$^\dagger$ \quad
  Nusrat Jahan Beg$^\dagger$ \quad
  Umme Hunny Khan \\
  \textbf{Syed Rifat Raiyan}$^\ddagger$ \quad
  \textbf{Md Kamrul Hasan} \quad
  \textbf{Hasan Mahmud} \\
  Systems and Software Lab (SSL), Department of Computer Science and Engineering \\
  Islamic University of Technology, Dhaka, Bangladesh \\
  \texttt{\small\{masnunnuha, nusratjahan21, ummehunny, rifatraiyan, hasank, hasan\}@iut-dhaka.edu} \\
  \small{$^\dagger$Equal contribution \quad $^\ddagger$Corresponding author}
}

\newcommand{\sysname}{\textsc{PROClaim}\xspace}

\newcommand{\micon}[1]{%
  \raisebox{-0.1\height}{\includegraphics[height=1.8ex]{images/#1.pdf}}%
  \hspace{0.4em}%
}

\newcommand{\GPTI}{\micon{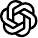}}
\newcommand{\DeepSeekI}{\micon{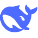}}

\newcommand{\QwenI}{\micon{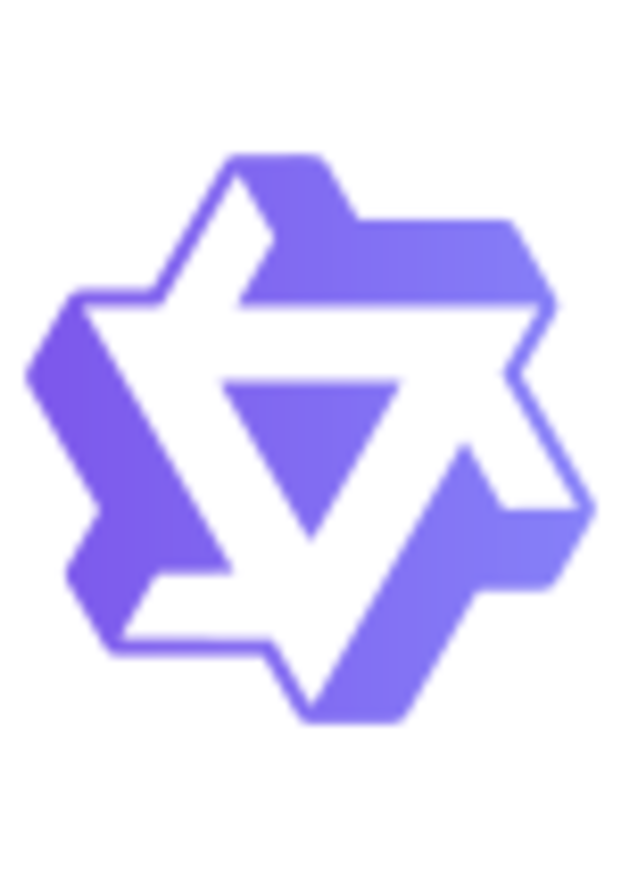}}

\newcommand{\LlamaI}{\micon{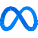}}

\newcommand{\squishlist}{%
  \begin{list}{$\bullet$}{%
    \setlength{\itemsep}{0pt}%
    \setlength{\parsep}{3pt}%
    \setlength{\topsep}{3pt}%
    \setlength{\partopsep}{0pt}%
    \setlength{\leftmargin}{1.5em}%
    \setlength{\labelwidth}{1em}%
    \setlength{\labelsep}{0.5em}%
  }%
}
\newcounter{Lcount}
\newcommand{\squishlisttwo}{%
  \begin{list}{\arabic{Lcount}.}{%
    \usecounter{Lcount}%
    \setlength{\itemsep}{0pt}%
    \setlength{\parsep}{0pt}%
    \setlength{\topsep}{0pt}%
    \setlength{\partopsep}{0pt}%
    \setlength{\leftmargin}{2em}%
    \setlength{\labelwidth}{1.5em}%
    \setlength{\labelsep}{0.5em}%
  }%
}
\newcommand{\squishend}{\end{list}}

\begin{document}
\maketitle
\begin{abstract}
Large language models (LLMs) remain unreliable for high-stakes claim verification due to hallucinations and shallow reasoning. While retrieval-augmented generation (RAG) and multi-agent debate (MAD) address this, they are limited by one-pass retrieval and unstructured debate dynamics. We propose a courtroom-style multi-agent framework, \sysname, that reformulates verification as a structured, adversarial deliberation. Our approach integrates specialized roles (\textit{e.g.}, Plaintiff, Defense, Judge) with Progressive RAG (P-RAG) to dynamically expand and refine the evidence pool during the debate. Furthermore, we employ evidence negotiation, self-reflection, and heterogeneous multi-judge aggregation to enforce calibration, robustness, and diversity. In zero-shot evaluations on the Check-COVID benchmark, \sysname achieves 81.7\% accuracy, outperforming standard multi-agent debate by 10.0 percentage points, with P-RAG driving the primary performance gains (+7.5 pp). We show that the majority of this improvement stems from P-RAG's dynamic coupling of retrieval to the evolving debate, rather than from any single component in isolation; the remaining courtroom mechanisms serve to stabilize and de-bias this process, together providing a robust foundation for reliable claim verification. 
\end{abstract}

\section{Introduction}
\label{sec:intro}
LLMs have demonstrated strong zero-shot performance on reasoning-intensive tasks, yet their reliability in high-stakes domains, such as claim verification, remains limited. Despite fluent generation, LLMs frequently exhibit hallucinations, shallow reasoning, and overconfident predictions when evaluating evidence-dependent claims \citep{Huang_2025}. RAG partially addresses these issues by grounding responses in external corpora \citep{lewis2020retrieval}. Still, standard pipelines rely on static, single-pass retrieval \citep{gao2024retrievalaugmentedgenerationlargelanguage} and lack mechanisms for iterative reasoning, often resulting in incomplete or biased conclusions.

To improve reliability, recent work has explored Multi-Agent Debate (MAD), where multiple LLM instances iteratively argue and refine answers \citep{du2024improving, liang2024encouraging, han2025debate}. While promising, prior studies show that unstructured debate often leads to premature convergence, shared bias reinforcement, and limited evidence exploration \citep{smit2023should, wu2025can}. In particular, agent agreement is frequently misinterpreted as correctness, even when grounded in insufficient or biased evidence.

In this work, we propose \sysname (\textbf{\underline{P}}rogressive \textbf{\underline{R}}etrieval \textbf{\underline{O}}rchestrated multi-agent framework for \textbf{\underline{Claim}} verification), centered on Progressive Retrieval-Augmented Generation (P-RAG), an iterative, query-adaptive retrieval mechanism that continuously expands and refines the evidence pool as debate unfolds. To generate the structured, evolving deliberation that P-RAG draws on for retrieval, we adopt a courtroom-style multi-agent framework that reformulates claim verification as a structured, adversarial reasoning process. Inspired by legal systems and recent agent-based simulations \citep{chen2025agentcourt}, this includes explicit roles (Plaintiff, Defense, Judge, Critic, and Expert Witness), evidence admission protocols, and multi-stage deliberation.\\
\textbf{Contributions.}
We present a unified framework where structured deliberation outperforms standard multi-agent debate, achieving a +10.0 pp accuracy gain and validating a courtroom-style architecture. Dynamic retrieval via P-RAG is the primary driver, adding 7.5 points while preventing evidence stagnation. We further show that model diversity is crucial: heterogeneous LLMs exhibit complementary errors that offset each other, outperforming any single model. \sysname also demonstrates strong zero-shot, domain-agnostic performance. Beyond gains, our analysis reveals deeper insights into multi-agent reasoning:
\squishlist
    \item \textbf{Breaking the epistemic bubble}, removing dynamic retrieval (P-RAG) \textbf{increases inter-judge agreement} ($\bar{\kappa}=0.468 \rightarrow 0.599$) while \textbf{reducing accuracy by 7.5 pp}, revealing confident convergence on incorrect conclusions.
    \item Acting as a \textbf{logic-level lie detector}, incorrect predictions exhibit \textbf{unstable reasoning trajectories} with oscillating self-reflection scores, showing that \textbf{reasoning dynamics are more informative than final confidence}.
    \item \textbf{Heterogeneous judges exhibit complementary error profiles} (\textit{e.g.}, over-refutation vs.\ cautious abstention), whose combination yields a \textbf{3.3 pp accuracy gain} over single-judge panels.
    \item Serving as an \textbf{economic governor}, removing self-reflection \textbf{increases debate rounds by 29\%} (5.47 $\rightarrow$ 7.06) and \textbf{token usage by 17\%} (210.9K $\rightarrow$ 247.3K), while changing accuracy by only \textbf{0.8 pp}.
    \item Demonstrating the \textbf{butterfly effect of argument framing}: small biases in premise decomposition \textbf{propagate to final outcomes}, as Run-2 reduces $\kappa_{GT}$ from 0.423(Run-0) to 0.384 despite stable inter-judge $\kappa$.
    \item Revealing \textbf{structural negativity bias}: LLM judges overproduce \textsc{Refute} relative to ground truth and converge faster on such claims (0.2 primary and 0.3 role-switched rounds), indicating training-induced conservatism.

\squishend

Together, these findings reframe LLM reasoning as a dynamic process, where reliability arises from the structure, diversity, and evolution of deliberation, not just final answers. We provide our code and data in the following GitHub repository: \url{https://github.com/mnc13/PROClaim}.

\section{Related Work}
\label{sec:related_short}
\textbf{Retrieval.} RAG grounds generation in external corpora \citep{lewis2020retrieval,
gao2024retrievalaugmentedgenerationlargelanguage}, and iterative variants improve coverage by
re-querying from intermediate generations \citep{shao-etal-2023-enhancing,
trivedi2023interleavingretrievalchainofthoughtreasoning,
park2025prograghallucinationresistantprogressiveretrieval}. Retrieval nonetheless remains driven by a
single reasoner's state, leaving it exposed to confirmatory bias under conflicting evidence
\citep{ge2025resolvingconflictingevidenceautomated}.

\textbf{Debate.} MAD improves factuality by having agents critique one another
\citep{du2024improving, liang2024encouraging} and has been adapted to fact-checking
\citep{han2025debate, ma2025local, he2025debating}. Controlled analyses, however, report premature
convergence, conformity bias, and sensitivity to configuration \citep{smit2023should, wu2025can,
zhu2026demystifying}: homogeneous panels rarely improve on majority vote. Hybrid systems attach
retrieval to debate \citep{hu2025removal, jeong2026tool, li2025r}, but draw from fixed or loosely
coupled evidence pools.

\textbf{Structure and coordination.} Courtroom simulations show that explicit roles stabilise
high-stakes deliberation \citep{chen2025agentcourt, chun2026agenticsimlaw, Jin2025CourtroomFND}, while
role assignment \citep{zhang2026dynamic}, role switching \citep{liu2025uncertainty}, self-reflection
\citep{madaan2023self, shinn2023reflexionlanguageagentsverbal}, dedicated critics
\citep{li2025headsbetteronedualmodel}, and multi-model juries
\citep{verga2024replacingjudgesjuriesevaluating} each address one failure mode in isolation.

\sysname couples them: retrieval queries are compiled from the live debate state and reflection gaps
rather than issued once upfront, role-switching is repurposed from bias mitigation into a consistency
diagnostic, judges evaluate the deliberative trajectory rather than a final answer, and termination is
governed by epistemic signals. Each ingredient exists somewhere; the closed loop, and the failure modes
it measurably prevents, are the contribution. Appendix~\ref{sec:related} expands this discussion.

\section{Methodology}
\label{sec:method}
We propose a courtroom-inspired pipeline for zero-shot, evidence-grounded fact-checking of COVID-19-related claims. Each claim is treated as a legal case: a \emph{Plaintiff Counsel} supports it, a \emph{Defense Counsel} challenges it, and an independent judicial panel delivers the verdict, imposing disciplined adversarial structure over the reasoning process.
Figure~\ref{fig:pipeline} provides a high-level overview, with the full evaluation cycle detailed in Appendix~\ref{pseudocode}.

\begin{figure*}[t]
    \centering
    \includegraphics[width=\linewidth]{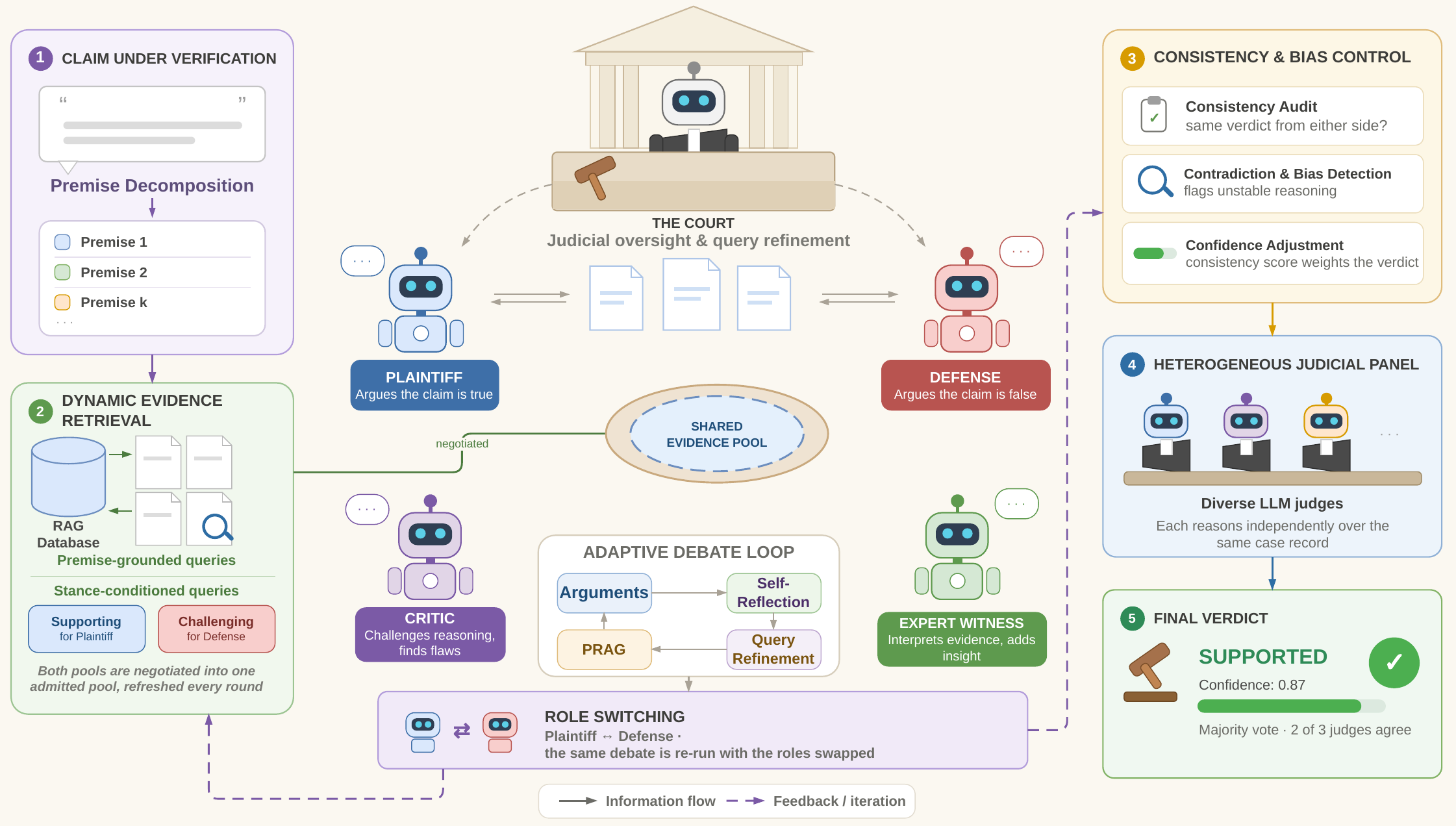}
    \caption{Overview of the courtroom-style claim-verification pipeline in \sysname.}
    \label{fig:pipeline}
    \vspace{-3mm}
\end{figure*}

\subsection{Argument Mining}
\label{sec:mining}

Before retrieval, the raw claim is decomposed into atomic, independently testable \emph{premises} \citep{hu2025decompositiondilemmasdoesclaim, lawrence-reed-2017-mining}. Given a claim $c$, the miner produces $\mathcal{P}=\{p_1,\ldots,p_k\}$, where each $p_i$ is a single verifiable proposition and the adaptive size $k$ captures the claim's non-redundant semantic content. These premises guide targeted retrieval and provide an explicit checklist for evaluating argument completeness during self-reflection and critic assessment (Section~\ref{sec:reflect}).

\subsection{Evidence Retrieval and Selection}
\label{sec:retrieval}

Relevant evidence is retrieved from a study-specific corpus of approximately 360,000 COVID-19 PubMed abstracts (2020--2024) using the \texttt{all-MiniLM-L6-v2} bi-encoder~\citep{gao2021condenserpretrainingarchitecturedense,reimers2019sentencebertsentenceembeddingsusing}. A FAISS index~\citep{douze2025faisslibrary} is queried once per premise (Section~\ref{sec:mining}) to form an initial shared pool while retaining publication and provenance metadata. Stance-conditioned supporting and opposing queries mitigate the bias of static Top-$k$ retrieval under conflicting evidence~\citep{wu2025biasinjectionattacksrag}; the resulting pools are exchanged before debate so each counsel can refine its case against the opposing evidence. Each retrieved document is assigned an admissibility score
\[
w=r\times c,
\]
where $r,c\in[0,1]$ denote relevance and credibility, respectively. Evidence with $w>0.5$ is admitted, $0.1<w\le0.5$ is marked as disputed, and the remainder is discarded. The admitted evidence is ranked by $w$ and passed to the debate stage (Appendix~\ref{app:admissibility}).

Our admissibility criterion follows the observation that evidence quality and provenance are as important as relevance when resolving conflicting claims~\citep{ge2025resolvingconflictingevidenceautomated}. Accordingly, we score each candidate jointly on relevance and credibility, inspired by the Daubert Standard~\citep{daubert_law_lII}, which emphasizes the admissibility of reliable scientific evidence.

\subsection{Progressive Retrieval-Augmented Generation (P-RAG)}
\label{sec:prag}

Static, one-time retrieval is ill-suited to adversarial debate as new evidential gaps emerge as arguments evolve. 
Unlike prior iterative approaches that query solely from the previous generation \citep{shao-etal-2023-enhancing, trivedi2023interleavingretrievalchainofthoughtreasoning}, P-RAG concatenates three sources: (i) the last four messages of the rolling debate context (a bounded recency window), (ii) the agent's self-identified evidential gap, and (iii) reflection-driven discovery needs from the prior round (Section~\ref{sec:reflect}), into one prompt with the Judge refining each query before execution (see App.~\ref{app:prag} for full logic and prompts). 
To prevent redundant retrieval, each candidate item is scored against the existing pool:

{
\begin{equation}
  \operatorname{novelty}(d) = 1 - \max_{p \in \mathcal{P}} \cos(e_d,\, e_p)
  \label{eq:novelty}
\end{equation}
}
where $\mathcal{P}$ is the current pool and $e_{\cdot}$ denotes an L2-normalised embedding. Unlike global diversity metrics such as the Vendi Score \citep{rezaei2025vendiragadaptivelytradingoffdiversity}, our novelty filter operates iteratively, rejecting near-duplicates at each round.

Only items with novelty $\geq 0.20$
are admitted. This threshold is not ad hoc: it was selected by grid search over $\{0.10, 0.15, 0.20, 0.25, 0.30\}$ on a held-out development subset; values below 0.15 allow near-duplicates, while values above 0.30 prematurely discard nuanced but critical evidence. Retrieval terminates early under the adaptive stopping criteria in Table~\ref{tab:prag}.

\begin{table}[t]
\centering
\small
\resizebox{\columnwidth}{!}{%
\begin{tabularx}{\columnwidth}{l p{1.3cm} X}
\hline
\textbf{Criterion} & \textbf{Threshold} & \textbf{Rationale} \\
\hline
Novelty filter   & \textcolor{blue!80!black}{$<0.20$} & Rejects near-duplicates \\
Redundancy ratio & \textcolor{blue!80!black}{$>70\%$} & Indicates saturation \\
Relevance gain   & \textcolor{blue!80!black}{$<0.05$} & Diminishing returns \\
Iteration cap    & \textcolor{purple!80!black}{$10$} & Limits compute cost \\
\hline
\end{tabularx}
}
\caption{P-RAG stopping criteria.}
\label{tab:prag}
\end{table}

\subsection{Multi-Agent Debate (MAD) Orchestration}
\label{sec:mad}

The multi-agent debate is the central reasoning engine of the framework, comprising five adjudication roles: Plaintiff Counsel, Defense Counsel, and three Judges. Each role is instantiated with an LLM appropriate for its function (Table~\ref{tab:models}); these assignments are illustrative rather than prescriptive, and the framework is compatible with alternative models. 
Controlled evidence indicates that homogeneous agents cannot reliably improve over majority voting~\citep{zhu2026demystifying}. We therefore introduce model heterogeneity across advocacy, critique, expert testimony, and adjudication; this complements dynamic role-assignment approaches~\citep{zhang2026dynamic} while promoting diverse reasoning and reducing correlated errors.

\begin{table}[t]
\centering
\small
\setlength{\tabcolsep}{3pt}
\resizebox{\columnwidth}{!}{%
\begin{tabular}{l l c c}
\hline
\textbf{Role} & \textbf{Model} & \textbf{Prov.} & \textbf{T.} \\
\hline
Premise Decomp.   & \DeepSeekI\textcolor{DeepSeekBlue}{\texttt{deepseek-r1}}       & OR & 0.7 \\
Plaintiff Counsel & \GPTI\textcolor{OpenAITeal}{\texttt{gpt-5-mini}}          & OA & 0.5 \\
Defense Counsel   & \DeepSeekI\textcolor{DeepSeekBlue}{\texttt{deepseek-v3.2}}     & OR & 0.5 \\
The Court         & \QwenI\textcolor{QwenIndigo}{\texttt{qwen3-235b-a22b}}          & OR & 0.2 \\
Expert Witness    & \LlamaI\textcolor{NousOrange}{\texttt{hermes-3-llama-405b}} & OR & 0.5 \\
Critic Agent      & \DeepSeekI\textcolor{DeepSeekBlue}{\texttt{deepseek-r1}}       & OR & 0.3 \\
Consistency Anal. & \DeepSeekI\textcolor{DeepSeekBlue}{\texttt{deepseek-v3.2}}     & OR & 0.3 \\
Judge 1           & \DeepSeekI\textcolor{DeepSeekBlue}{\texttt{deepseek-r1}}       & OR & 0.3 \\
Judge 2           & \LlamaI\textcolor{NousOrange}{\texttt{hermes-3-llama-405b}} & OR & 0.3 \\
Judge 3           & \QwenI\textcolor{QwenIndigo}{\texttt{qwen3-235b-a22b}}     & OR & 0.3 \\
\hline
\end{tabular}
}
\caption{Role-wise models. Providers (Prov.): \textbf{OR}=OpenRouter, \textbf{OA}=OpenAI. T.=Temperature.}
\label{tab:models}
\vspace{-3mm}
\end{table}

Each round consists of five stages: \textbf{(1) Evidence discovery}, where both agents identify evidential gaps and retrieve additional evidence via P-RAG; \textbf{(2) Argument generation}, producing evidence-grounded arguments; \textbf{(3) Expert testimony}, where the Judge may invoke a dynamically instantiated domain expert when needed; \textbf{(4) Self-reflection}, whose identified gaps are injected into the next P-RAG query (Section~\ref{sec:reflect}); and \textbf{(5) Critic evaluation}, where an independent Critic assesses both arguments and may trigger early termination.

The debate runs for at most 10 rounds and terminates early if any of the following conditions are met: (i) reflection plateau ($|\Delta\,\texttt{total\_reflection\_score}|<0.05$ for two consecutive rounds), (ii) critic resolution (\texttt{debate\_resolved=True}), (iii) novelty exhaustion (average novelty $<0.10$ across two consecutive P-RAG calls), or (iv) judicial signal indicating readiness for deliberation.

\subsection{Self-Reflection and Critic Evaluation}
\label{sec:reflect}

\textbf{Per-agent self-reflection.}
After each round, each counsel performs structured self-reflection~\citep{madaan2023self, shinn2023reflexionlanguageagentsverbal} along three dimensions: \emph{logic} ($l$), \emph{novelty} ($n$), and \emph{rebuttal} ($b$), each in $[0,1]$. Their relative weights (0.4, 0.3, 0.3) were determined \emph{a priori} through a structured expert elicitation following the SHELF/Delphi protocol (Appendix~\ref{app:elicitation}). The reflection score is computed as

\begin{equation}
  s_{\text{ref}} = 0.4\,l + 0.3\,n + 0.3\,b,
  \label{eq:reflect}
\end{equation}

and converted into a confidence adjustment,

\begin{equation}
  \delta_{\text{ref}} = (s_{\text{ref}} - 0.5) \times 0.6 \in [-0.30,\,+0.30],
  \label{eq:conf_adj}
\end{equation}

The $0.6$ multiplier limits the adjustment to $\pm0.30$, allowing self-reflection to refine without overriding the panel's baseline confidence ($c_{\text{base}}\approx0.8$). The reflection output also identifies missing evidence for the subsequent P-RAG retrieval query.

\textbf{Independent Critic Agent.}
An independent Critic Agent evaluates each round, motivated by prior work showing dedicated critics outperform self-critique~\citep{li2025headsbetteronedualmodel}. It scores both arguments on logic, evidence coverage, and rebuttal, identifies unresolved premises (Section~\ref{sec:mining}), provides feedback, and issues the \texttt{debate\_resolved} signal.

\subsection{Role-Switching Consistency Test}
\label{sec:roleswitch}
We introduce a \textbf{role-switching consistency test} to distinguish evidence-grounded reasoning from position anchoring. Unlike prior work using role-switching for bias reduction or uncertainty estimation \citep{Jin2025CourtroomFND, liu2025uncertainty}, we use it diagnostically. After the primary debate, the counsels swap roles, the state is reset, and the debate is repeated. A separate LLM compares both transcripts for contradictions and selective evidence use, producing an agent-level \emph{consistency score} for final confidence weighting (Section~\ref{sec:judges}).

\subsection{Judicial Panel Evaluation and Final Verdict}
\label{sec:judges}

The complete case record, including both debate transcripts, admitted evidence, P-RAG histories, critic evaluations, and self-reflection scores, is submitted to three independent LLM judges using different models to reduce correlated errors \citep{verga2024replacingjudgesjuriesevaluating}. Each judge evaluates case reconstruction, evidence strength, argument validity, source reliability, and retrieval rigour before assigning one of three verdicts: \textsc{Supported}, \textsc{Not Supported}, or \textsc{Inconclusive}. The final verdict is determined by majority vote (Appendix~\ref{app:consensus}).

\textbf{Base confidence:}
\begin{equation}
  c_{\text{base}} = 0.8\,\sigma + 0.3\,q,
  \label{eq:base_conf}
\end{equation}
where $\sigma=\texttt{winning\_votes}/\texttt{total\_votes}$ represents consensus strength and
$q = (\bar{s}_{\text{ev}} + \bar{s}_{\text{val}} + \bar{s}_{\text{rel}})/30$  is the normalized mean score for evidence, validity, and reliability. The consensus weight was selected through 5-fold cross-validated grid search to minimize Expected Calibration Error \citep{gu2025surveyllmasajudge}. The final configuration achieved an ECE of $0.034$, compared with $0.18$ for standard averaging (Appendix~\ref{app:calibration}; weight details in Appendix~\ref{app:appendix_confidence_weights}).

\textbf{Adjustment:}
\begin{equation}
  c_{\text{final}} = \text{clamp}\bigl(c_{\text{base}} + \delta_{\text{rs}} + \delta_{\text{ref}},\;0,\;1\bigr),
  \label{eq:final_conf}
\end{equation}
where $\delta_{\text{rs}}$ is the role-switch adjustment and $\delta_{\text{ref}}$ is the winning side's final reflection adjustment. The latter is lower-bounded at $-0.15$ to prevent self-critique from overriding the judicial consensus. When at least two judges agree, the final confidence is also lower-bounded at $0.10$ (Appendices~\ref{app:appendix_roleswitch} and \ref{app:appendix_reflection}).

\section{Experimental Setup}
\label{sec:exp}

\textbf{Dataset.} To evaluate the framework's capacity for adversarial resolution, we focus on the subset of the Check-COVID \citep{wang2023checkcovidfactcheckingcovid19news} test set possessing definitive binary ground-truths (SUPPORT or REFUTE), comprising 120 claims evaluated across three independent runs (360 debate instances). \par

\textbf{Retrieval Corpus.} Retrieval is performed over a static corpus of COVID-19 abstracts from PubMed (2020--2024), constructed for this study. Source metadata, including PMID, title, journal, and publication year, are stored alongside each passage to support provenance tracking.\par
\textbf{Configuration and Metrics.} Table~\ref{tab:models} summarizes the LLM assignment, where agent roles are partitioned across distinct architectures based on their functional capacity (deliberative \textit{vs.}\ evaluative) and cognitive profile, utilizing moderate temperatures for adversarial advocacy and lower temperatures for rigorous judicial evaluation (details in Appendix~\ref{app:appendix_role_separation}).  PROClaim is model-agnostic by design; the reported configuration is one
instantiation. Key retrieval hyperparameters are a novelty threshold $\tau{=}0.20$, per-round top-$k{=}3$, and admissibility floor $w{>}0.5$; full settings are in Appendix~\ref{app:hyperparams}.

We evaluate classification performance (accuracy, macro F1), inter-judge reliability (Cohen's $\kappa$, unanimity/split rates), and efficiency (token usage, debate rounds, evidence pool size).

\section{Results and Discussion}
\label{sec:results}
Table~\ref{tab:framework_results} reports Check-COVID performance across three independent runs, with aggregate results obtained via majority voting. Our framework adopts an adversarial \emph{Burden of Refutation}: if a claim withstands deliberation and the judicial panel cannot reach a definitive \textsc{Refute} verdict, returning \textsc{Inconclusive} due to insufficient opposing evidence, it is classified as \textsc{Supported}. This evaluation protocol follows the courtroom principle that claims are upheld unless successfully refuted (Appendix~\ref{app:appendix_burden}).

\begin{table}[t]
\centering
\small
\setlength{\tabcolsep}{3.5pt}

\resizebox{\columnwidth}{!}{%
\begin{tabular}{l c c c c c c }
\hline
\textbf{Run} &
\textcolor{blue!80!black}{\textbf{Acc}} &
\textcolor{blue!80!black}{\textbf{m-F1}} &
\textcolor{purple!80!black}{\textbf{$\bar{\kappa}$}} &
\textcolor{purple!80!black}{\textbf{Agr.}} &
\textcolor{purple!80!black}{\textbf{Unan.}} &
\textcolor{purple!80!black}{\textbf{Split}} \\
\hline

Run-0     & 0.950 & 0.950 & 0.429 & 0.617 & 0.442 & 0.558 \\

Run-1     & 0.817 & 0.817 & 0.549 & 0.700 & 0.558 & 0.442 \\

Run-2     & 0.790 & 0.790 & 0.474 & 0.652 & 0.496 & 0.504 \\

\hline
W. Total  & 0.841 & 0.841 & 0.484 & 0.656 & 0.499 & 0.501 \\

\rowcolor{gray!15}
Maj. Vote & \textbf{0.817} & \textbf{0.817} & \textbf{0.468} & \textbf{0.648} & \textbf{0.489} & \textbf{0.511} \\

\textcolor{gray!70!black}{\textit{Oracle}} &
0.958 & 0.958 & 0.438 & 0.622 & 0.450 & 0.550 \\
\hline
\end{tabular}
}

\caption{Performance statistics. \textbf{m-F1}: MacroF1. \textbf{$\bar{\kappa}$}: Mean $\kappa$. \textbf{Agr.}: Raw Agreement. \textbf{Unan.}: Unanimity (3--0 consensus). \textbf{Split}: Divided vote. \textbf{W. Total}: Weighted Total. \textbf{Oracle}: Best-of-3. Per-judge \textbf{$\kappa_\text{GT}$} in App.~\ref{app:kgt}.}
\label{tab:framework_results}
\vspace{-3mm}

\end{table}

\subsection{Main Pipeline Performance} \label{sec:main_results} Majority voting achieves 81.7\% accuracy, while the 95.8\% oracle ceiling bounds performance under favourable initialization. Inter-judge agreement remains stable across runs ($\kappa$: 0.429, 0.549, 0.474) despite expected variance in cascaded LLM pipelines~\citep{atil2025non}, which \sysname's audit trail makes interpretable. Run-1 retains 0.817 accuracy despite lower ground-truth agreement ($\kappa_{GT}=0.372$) through balanced errors. In Run-2, biased premise decomposition suppresses \textsc{Refute} recall and lowers $\kappa_{GT}$ to 0.384 while inter-judge $\kappa$ remains 0.474, indicating systematic bias rather than structural failure.

\subsection{Debate Dynamics and Adaptive Stopping}
\label{sec:dynamics}

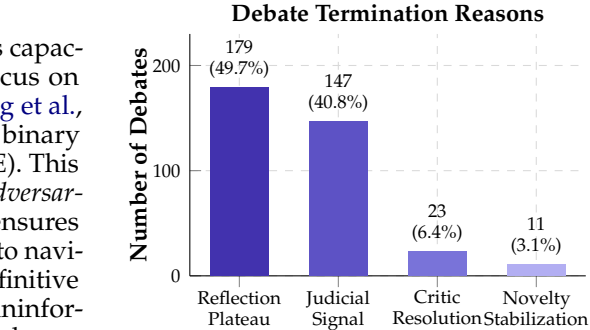
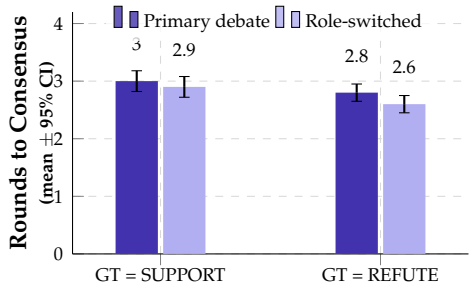
\begin{figure}[t]
\centering

\definecolor{p1}{HTML}{B2AEF2} 
\definecolor{p2}{HTML}{9491E4}
\definecolor{p3}{HTML}{7973D9}
\definecolor{p4}{HTML}{5D52C5}
\definecolor{p5}{HTML}{4232AE} 

\begin{subfigure}{\linewidth}
\centering
\begin{tikzpicture}
\begin{axis}[
    width=\linewidth,
    height=4.8cm,
    ybar,
    bar width=22pt,
    bar shift=0pt,
    ymin=0, ymax=230,
    xmin=0.5, xmax=4.5,
    axis y line*=left,
    axis x line*=bottom,
    ylabel={Number of Debates},
    ylabel style={font=\footnotesize\bfseries, yshift=-1ex},
    y tick label style={font=\scriptsize},
    xtick={1,2,3,4},
    xticklabels={Reflection\\Plateau, Judicial\\Signal, Critic\\Resolution, Novelty\\Stabilization},
    x tick label style={align=center, font=\scriptsize, text width=1.6cm, yshift=0.5ex},
    grid=major,
    grid style={dashed, gray!30},
    title={\textbf{Debate Termination Reasons}},
    title style={font=\footnotesize, yshift=-1ex}
]

\addplot[fill=p5, draw=none] coordinates {(1, 179)};
\node[align=center, font=\scriptsize, above] at (axis cs:1,179) {179\\(49.7\%)};

\addplot[fill=p4, draw=none] coordinates {(2,147)};
\node[align=center, font=\scriptsize, above] at (axis cs:2,147) {147\\(40.8\%)};

\addplot[fill=p3, draw=none] coordinates {(3,23)};
\node[align=center, font=\scriptsize, above] at (axis cs:3,23) {23\\(6.4\%)};

\addplot[fill=p1, draw=none] coordinates {(4,11)};
\node[align=center, font=\scriptsize, above] at (axis cs:4,11) {11\\(3.1\%)};

\end{axis}
\end{tikzpicture}
\caption{Debate termination reasons.}
\label{fig:debate_termination}
\end{subfigure}
\hfill
\begin{subfigure}{\linewidth}
\centering
\begin{tikzpicture}
\begin{axis}[
    width=\linewidth,
    height=4.8cm,
    ybar=2pt,
    bar width=16pt,
    ymin=0, ymax=4.2,
    axis y line*=left,
    axis x line*=bottom,
    ylabel={\begin{tabular}{c}Rounds to Consensus\\ \scriptsize(mean $\pm$ 95\% CI)\end{tabular}},
    ylabel style={align=center, font=\footnotesize\bfseries, yshift=-2ex},
    y tick label style={font=\scriptsize},
    symbolic x coords={GT = SUPPORT, GT = REFUTE},
    xtick=data,
    x tick label style={font=\scriptsize, yshift=0.5ex},
    grid=major,
    grid style={dashed, gray!30},
    enlarge x limits=0.4,
    legend style={
        at={(0.5,1.05)},
        anchor=north,
        legend columns=-1,
        draw=none,
        fill=white,
        fill opacity=0.8,
        text opacity=1,
        font=\scriptsize
    },
    nodes near coords={\pgfmathprintnumber[fixed,precision=1]{\pgfplotspointmeta}},
    every node near coord/.append style={font=\scriptsize,yshift=8pt,color=black}
]

\addplot[
    fill=p5,
    draw=none,
    error bars/.cd,
    y dir=both,
    y explicit,
    error bar style={color=black, thick}
] coordinates {
    (GT = SUPPORT,3.0) +- (0,0.18)
    (GT = REFUTE,2.8) +- (0,0.15)
};

\addplot[
    fill=p1,
    draw=none,
    error bars/.cd,
    y dir=both,
    y explicit,
    error bar style={color=black, thick}
] coordinates {
    (GT = SUPPORT,2.9) +- (0,0.18)
    (GT = REFUTE,2.6) +- (0,0.15)
};

\legend{Primary debate, Role-switched}

\end{axis}
\end{tikzpicture}
\caption{Rounds to consensus by ground-truth label.}
\label{fig:rounds_consensus}
\end{subfigure}

\caption{Termination distribution and convergence speed across 360 debate instances.}
\label{fig:debate_stats}
\vspace{-3mm}
\end{figure}

\begin{figure*}[t]
    \centering

    \begin{subfigure}[t]{0.45\textwidth}
        \centering
        \includegraphics[width=\linewidth]{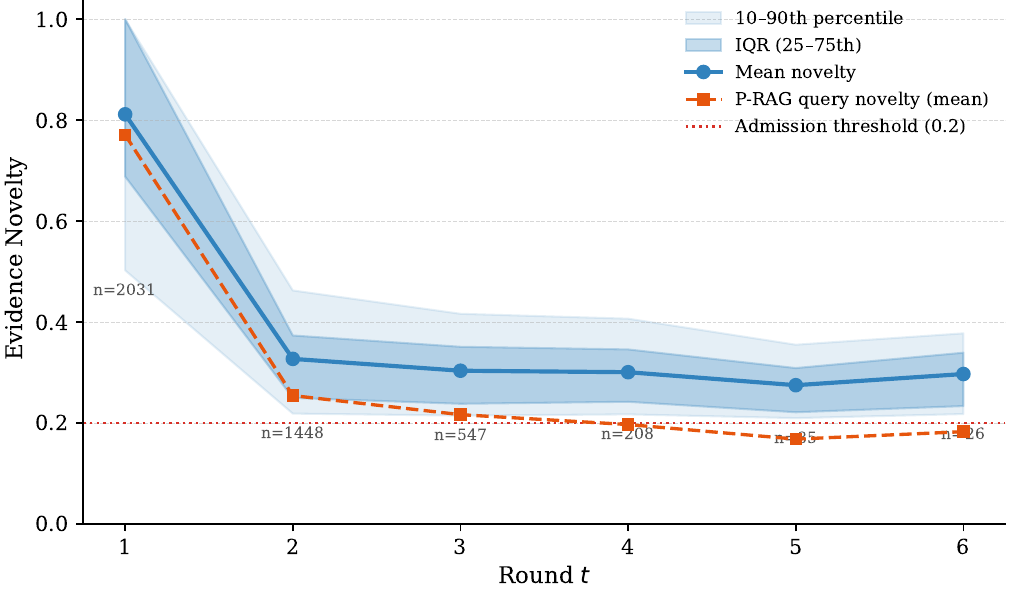}
        \caption{Novelty decay across rounds.}
        \label{fig:novelty_decay}
    \end{subfigure}
    \hfill
    \begin{subfigure}[t]{0.45\textwidth}
        \centering
        \includegraphics[width=\linewidth]{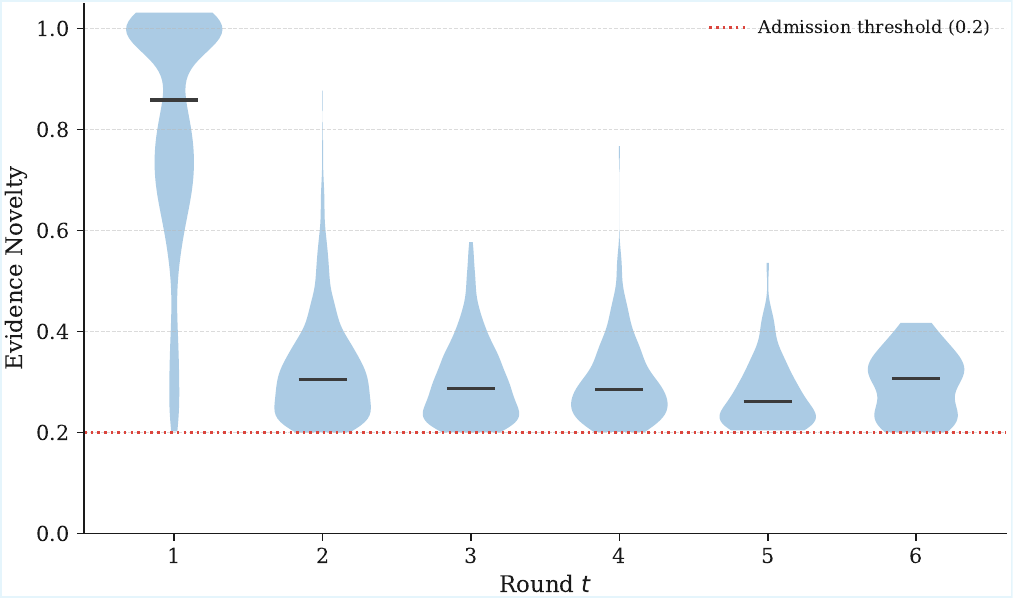}
        \caption{Novelty distribution by round.}
        \label{fig:novelty_violin}
    \end{subfigure}

    \caption{P-RAG evidence novelty across debate rounds.}
    \label{fig:novelty}
    \vspace{-3mm}
\end{figure*}

\begin{figure*}[t]
    \centering
    \includegraphics[width=0.9\textwidth]{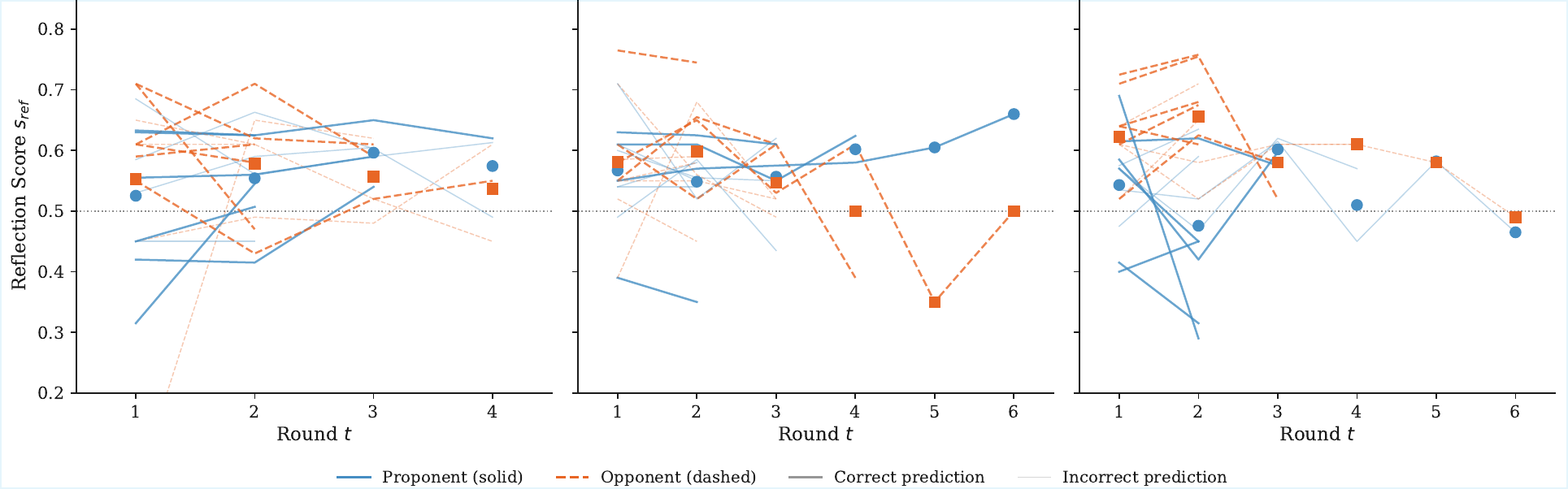}
    \caption{Reflection score trajectories across plateau, judicial, and critic resolution patterns.}
    \label{fig:reflection}
    \vspace{-3mm}
\end{figure*}

\textbf{Termination distribution: }
Figure~\ref{fig:debate_termination} shows that most debates self-terminate through reflection-driven signals: Reflection Plateau and Judicial Signal. Reflection Plateau enforces adaptive halting once marginal argumentative gains diminish, mitigating circular reasoning. The low incidence of Novelty Stabilization (3.1\%) further suggests that termination is driven by argumentative saturation rather than corpus insufficiency.

\textbf{Convergence speed: }
Figure~\ref{fig:rounds_consensus} shows that \textbf{REFUTE claims converge faster than SUPPORT claims} in both the primary (0.2 rounds faster) and role-switched debates (0.3 rounds faster), consistent with the LLM negativity bias documented in Section~\ref{sec:judgebias}.

\textbf{Evidence novelty decay: }
Figures~\ref{fig:novelty_decay}--\ref{fig:novelty_violin} confirm that P-RAG sustains \textbf{genuine evidential discovery across all active rounds}, with query novelty approaching the threshold from round~4 onward. The narrowing violin distributions from round~3 indicate pool saturation, \textbf{empirically validating the 0.20 novelty threshold as the natural boundary between productive and redundant retrieval}, justifying the adaptive stopping design.

\textbf{Reflection trajectories: }
Figure~\ref{fig:reflection} plots per-claim self-reflection score trajectories across debate rounds $t$: each line corresponds to one claim-level debate ($\sim$10 representative claims sampled per panel across runs), and each panel stratifies debates by their termination condition. The stratification reveals distinct convergence behaviours: Reflection Plateau yields rapid stabilisation within 3--4 rounds, Judicial Signal extends deliberation for contested claims, and Critic Resolution exhibits higher variance. Across all conditions, incorrect predictions display greater inter-round oscillation whereas correct predictions converge smoothly, a behavioural signature of erroneous outcomes that single-pass pipelines cannot expose.

\subsection{Comparison with Baselines}
\label{sec:baselines}

\begin{table}[t]
\centering
\small
\setlength{\tabcolsep}{3pt}
\resizebox{\columnwidth}{!}{%
\begin{tabular}{lccc}
\hline
\textbf{System} &
\textcolor{blue!80!black}{\textbf{Acc}} &
\textcolor{blue!80!black}{\textbf{MacroF1}} &
\textcolor{purple!80!black}{\textbf{Ev}} \\
\hline

Single-call \textcolor{DeepSeekBlue}{DeepSeek-v3.2} + RAG &
0.8000 & 0.7972 & 18.5 \\

Standard MAD &
0.7167 & 0.7068 & 12.1 \\
\hline

\rowcolor{gray!15}
\textbf{\sysname\ (Majority Vote)} &
\textbf{0.8167} &
\textbf{0.8165} &
\textbf{67.5} \\
\hline

\end{tabular}%
}

\caption{\sysname\ \textit{vs.}\ baselines on Check-COVID. Ev denotes average evidence pool size per claim.}
\label{tab:baseline}
\vspace{-3mm}
\end{table}

Against two-agent, single-judge Standard MAD (71.7\%), \sysname's gain is significant on HealthVer (McNemar's $\chi^2=8.04$, $p=0.0046$, odds ratio 3.67; Section~\ref{sec:external}).


Single-call RAG is competitive in-domain (80.0\%) but framing-sensitive: identical evidence supports opposing arguments in our role-switch case study (\autoref{app:roleswitching}). On HealthVer it falls to 63.0\%, while \sysname declines from 81.7\% to 72.0\%, widening the gap from 1.7 to 9.0 pp (Section~\ref{sec:external}). Single-agent iterative RAG would confound retrieval policy with architecture; our controlled comparison is the w/o-P-RAG ablation ($-7.5$ pp, Section~\ref{sec:ablation}).

Beyond accuracy, \sysname prioritizes \textbf{Deliberative Traceability}: variance in judge scores and agent consistency surfaces a Trajectory Instability Signal, a ``Logic Lie Detector'' for contested claims inherently absent in single-call black-box pipelines (artifact inventory in Appendix~\ref{app:traceability}), and its confidence is nearly perfectly calibrated (ECE $=0.034$ \textit{vs.}\ 0.18 under naive aggregation; Appendix~\ref{app:calibration}). The 95.8\% Oracle ceiling further establishes that the architecture's attainable bound exceeds every reported baseline.

\subsection{Ablation Study}
\label{sec:ablation}
We conduct four ablation experiments, each excluding a single subsystem, evaluated on the same 120 claims.  Table~\ref{tab:ablation_structure} summarises component activation across all configurations.

\begin{table}[t]
\centering
\small
\setlength{\tabcolsep}{3pt}
\newcommand{\cmark}{\textcolor{green!60!black}{\textbf{\checkmark}}}
\newcommand{\xmark}{\textcolor{red}{$\mathbf{\times}$}}
\newcommand{\pmark}{\textcolor{orange!80!black}{Partial}}

\resizebox{0.95\columnwidth}{!}{%
\begin{tabular}{lccccc}
\hline
\textbf{Component} & \textbf{Full} & \textbf{A1} & \textbf{A2} & \textbf{A3} & \textbf{A4} \\
\hline

Evidence Negotiation   & \cmark & \cmark & \cmark & \cmark & \cmark \\
P-RAG                  & \cmark & \cmark & \cmark & \xmark & \cmark \\
Expert Witnesses       & \cmark & \cmark & \cmark & \cmark & \cmark \\
Self-Reflection        & \cmark & \cmark & \cmark & \cmark & \xmark \\
Reflection $\to$ P-RAG & \cmark & \cmark & \cmark & \xmark & \xmark \\
Critic Agent           & \cmark & \cmark & \cmark & \cmark & \cmark \\
Adaptive Convergence   & \cmark & \cmark & \cmark & \xmark & \pmark \\
Role-Switching         & \cmark & \xmark & \cmark & \cmark & \cmark \\
3-Judge Panel          & \cmark & \cmark & \xmark & \cmark & \cmark \\
Reflection Confidence Adj. & \cmark & \cmark & \cmark & \cmark & \xmark \\
\hline

Max Rounds             & 10 & 10 & 10 & 3 & 10 \\
\hline
\end{tabular}
}

\caption{Component ablation matrix. \cmark~= active; \xmark~= disabled. {\small \textit{Note:} A1: no role-switch; A2: 1-judge; A3: no P-RAG; A4: no self-reflection.}}
\label{tab:ablation_structure}
\vspace{-5mm}
\end{table}

\begin{table*}[t]
\centering
\setlength{\tabcolsep}{5pt} 
\resizebox{0.90\textwidth}{!}{%
\begin{tabular}{@{} l c c c c c c c @{}}
\hline
\textbf{System} & \textbf{Acc} & \textbf{MacroF1} & $\boldsymbol{\Delta}$\textbf{Acc} & {\textbf{$\bar{\kappa}$}} & \textbf{Avg. Rounds} & \textbf{Ev} & \textbf{Tok (K)} \\
\hline
\rowcolor{gray!15} \textbf{\sysname\ (Ours)} & \textbf{0.8167} & \textbf{0.8165} & --- & 0.468 & 5.47 & 67.5 & 210.9 \\
\hline
w/o P-RAG             & 0.7417 & 0.7408 & \textbf{$-$7.5 pp} & 0.599 & 6.00 & 37.5 & 188.9 \\
w/o Role-Switching    & 0.7750 & 0.7750 & $-$4.2 pp & 0.513 & 2.88 & 54.0 & 147.3 \\
w/o Three-Judge Panel & 0.7833 & 0.7818 & $-$3.3 pp & --    & 5.29 & 68.8 & 195.9 \\
w/o Self-Reflection   & 0.8083 & 0.8080 & $-$0.8 pp & 0.591 & 7.06 & 81.5 & 247.3 \\
\hline
Standard MAD          & 0.7167 & 0.7068 & \textbf{$-$10.0 pp}& --    & 2.00 & 12.1 & 18.9 \\
\hline
\end{tabular}
}
\caption{Ablation results on 120 claims. $\Delta$Acc is measured relative to the full \sysname\ pipeline. Ev denotes the average evidence pool size per claim. Each round count reflects the sum of primary and role-switched debate rounds per claim.}
\label{tab:ablation}
\vspace{-5mm}
\end{table*}
\textbf{P-RAG is the most impactful component} ($\downarrow 7.5$ pp): without it, the evidence pool nearly halves (37.5 \textit{vs.}\ 67.5), debates run longer (6.00 \textit{vs.}\ 5.47) on weaker evidence, and inter-judge agreement rises to $\kappa=0.599$, a \emph{negative} signal of convergence without correctness (\emph{epistemic bubble}). 
\textbf{Role-switching contributes $-4.2$ pp} and reduces the evidence pool by 20\%, confirming the second pass surfaces missed evidence rather than serving as a mere consistency check.  
\textbf{A single judge costs 3.3 pp}, directly quantifying the benefit of heterogeneous adjudication (Sec.~\ref{sec:judges}). 
\textbf{Self-reflection has minimal accuracy impact ($-0.8$ pp) but drives efficiency}: removing it increases rounds by 29\% (5.47 $\rightarrow$ 7.06) and token usage by 17\% (210.9K $\rightarrow$ 247.3K), confirming that its early-stopping signal makes the pipeline cost-effective. A full token usage breakdown and analysis are provided in Appendix~\ref{app:cost}.

\subsection{Judicial Panel: Reliability and Negativity Bias}
\label{sec:judgebias}

Inter-judge agreement ($\kappa_\text{mean}=0.468$; 48.9\% unanimous) shows that the heterogeneous panel disagrees in 51.1\% of cases. DeepSeek-R1 \emph{aggressively refutes} (high \textsc{Refute} recall but false-refutes \textsc{Support}), Hermes-3-405B \emph{cautiously abstains} (frequent \textsc{Inconclusive} on \textsc{Support}), and Qwen3-235B-A22B remains the \emph{most calibrated} (highest \textsc{Support} recall and balanced abstention). Although all overproduce \textsc{Refute} or \textsc{Inconclusive}, indicating structural negativity bias, their distinct commission, omission, and calibration errors allow disagreement to correct rather than compound mistakes. Accordingly, majority voting achieves balanced judge--ground-truth agreement ($\kappa_\text{GT}\in[0.431,0.468]$, mean $0.450$), supporting heterogeneous voting as a means of improving reliability.

\subsection{Cross-Domain Generalization and Paired Baseline Comparison}
\label{sec:external}

\begin{table}[b]
\centering
\setlength{\tabcolsep}{5pt} 
\resizebox{\columnwidth}{!}{%
\begin{tabular}{lcccc}
\hline
\textbf{Dataset} & \textbf{Claim} &
\textcolor{blue!80!black}{\textbf{Acc}} &
\textcolor{blue!80!black}{\textbf{F1}} &
\textcolor{purple!80!black}{\textbf{Tok(K)}} \\
\hline
HealthVer & 100 & 0.720 & 0.713 & 223 \\
FEVEROUS  &  60 & 0.783 & 0.772 & 236 \\
\hline
\rowcolor{gray!15}
Check-COVID & 120 & 0.817 & 0.817 & 211 \\
\hline
\end{tabular}
}
\caption{Generalization results (single run)}
\label{tab:external}
\end{table}

\textbf{HealthVer.} On a uniformly sampled set of 100 claims, the framework achieves 72.0\% accuracy \citep{sarrouti-etal-2021-evidence-based}.\
\textbf{FEVEROUS.} Using a Wikipedia-based retriever, it achieves 78.3\% accuracy on 60 sampled claims, within $\sim$3 pp of Check-COVID \citep{Aly21Feverous}.

These proof-of-concept robustness checks use sampled subsets because full-corpus deliberation requires approximately 211K tokens per claim. With zero domain-specific tuning and only the retrieval backend changed, accuracy remains within 9.7 pp across datasets, supporting the generality of the deliberative architecture (Appendix~\ref{sec:external_datasets}).

\begin{table}[ht]
\centering
\setlength{\tabcolsep}{6pt}
\resizebox{\columnwidth}{!}{%
\begin{tabular}{@{} l c c c @{}}
\hline
\textbf{System} &
\textcolor{blue!80!black}{\textbf{Acc}} &
\textcolor{blue!80!black}{$\boldsymbol{\Delta}$ \textbf{(pp)}} &
\textcolor{purple!80!black}{\textbf{Odds Ratio}} \\
\hline
Single-call \textcolor{DeepSeekBlue}{DeepSeek-v3.2} + RAG & 63.0\% & $+9.0$  & 2.50 \\
Standard MAD                                              & 56.0\% & $+16.0$ & \textbf{3.67}$^{\ddagger}$ \\
\midrule
\rowcolor{gray!15}
\textbf{\sysname} & \textbf{72.0\%} & --- & --- \\
\hline
\end{tabular}
}
\caption{Paired comparison on the identical 100-claim HealthVer subset. $\Delta$ is \sysname's accuracy advantage; the odds ratio $\mathrm{OR}=b/c$ counts claims \sysname resolves correctly where the baseline fails ($b$) against the converse ($c$), so $\mathrm{OR}>1$ means \sysname wins the contested cases. $^{\ddagger}$Statistically significant (McNemar's $\chi^2=8.04$, $p=0.0046$). Protocol details in Appendix~\ref{app:stats}.}
\label{tab:healthver_paired}
\vspace{-5mm}
\end{table}

\textbf{Paired baseline comparison.} We re-evaluate the baselines on the \emph{identical} 100-claim HealthVer subset, enabling valid paired analysis (two-sided continuity-corrected McNemar's tests, bootstrap $n=10{,}000$; Appendix~\ref{app:stats}). Table~\ref{tab:healthver_paired}: \sysname outperforms every baseline, with odds ratios uniformly above 1 and a statistically significant margin over Standard MAD ($p=0.0046$) mirroring the +10.0 pp Check-COVID advantage. The comparison also exposes single-model RAG brittleness; DeepSeek-v3.2+RAG falls from 80.0\% in-domain to 63.0\% off-domain, a 17 pp collapse against \sysname's 9.7 pp: parametric shortcuts do not generalize; the deliberative architecture's gains persist.

\subsection{Sycophancy and Argument Integrity}

\label{sec:sycophancy}
Sycophancy, the tendency of agents to yield to opposing arguments \citep{malmqvist2025sycophancy}, is low overall: \textbf{role-play consistency} (Plaintiff 7.80/10; Defense 6.22/10) shows only mild asymmetry from Defense ``leaking'' prior-role reasoning after switches; \textbf{concession rates} are minimal (0.03 / 0.01 per 1{,}000 words), the higher Plaintiff rate consistent with the negativity bias (Section~\ref{sec:judgebias}); the \textbf{reflection plateau} averages 0.3823 per round, far above the 0.05 stagnation threshold, confirming agents push novel reasoning rather than colluding; and \textbf{judicial conformity} (Fleiss' $\kappa = 0.4513$) affirms independent evaluation while precluding the absolute conformity ($\kappa \to 1.0$) symptomatic of systemic sycophancy (Appendix~\ref{app:sycophancy}).

\section{Conclusion}
\label{sec:conclusion}
\sysname{} shows that courtroom-style deliberation yields a claim-verification system that is accurate, calibrated, and auditable. By closing the loop between debate and retrieval, and policing it with procedural safeguards borrowed from legal practice, we establish that reliability in multi-agent LLM systems emerges from deliberative architecture rather than individual model capability, offering a principled, empirically validated template for high-stakes verification. Future work includes live PubMed retrieval, extension to other high-stakes domains, and token reduction via early-exit and distillation.

\section*{Limitations}
\label{sec:limitations}
\textbf{Computational cost.} Full deliberation consumes $\sim$211K tokens per claim, roughly $11\times$ Standard MAD.  Our ablations show each costly component buys measurable accuracy; Appendix~\ref{app:cost} maps the cost--accuracy Pareto frontier with reduction pathways for latency-sensitive settings.

\textbf{Run-level variance.} Identically configured runs vary (0.790--0.950 accuracy), consistent with non-determinism in cascaded LLM pipelines \citep{atil2025non}. Our audit record makes this interpretable (Section~\ref{sec:main_results}); majority voting bounds but does not eliminate it.

\textbf{Rationale faithfulness.} The structured case record (evidence with provenance, transcripts, reflection/consistency scores) is computed programmatically and auditable; judges' \emph{verbal} rationales, however, inherit the open problem that LLM chains-of-thought need not reflect underlying computation \citep{chen2025reasoningmodelsdontsay}. \sysname{} thus offers \emph{deliberative traceability}, complementary to and weaker than the formal verifiability of argumentation-theoretic frameworks \citep{Freedman_2025}; we do not claim faithful explanation.


\textbf{Evaluation scale and instantiation.} Cross-domain results use sampled subsets (100 HealthVer / 60 FEVEROUS, full-corpus deliberation being cost-prohibitive) and are proof-of-concept checks with paired significance testing. The reported role assignment is one budget-constrained instantiation of a model-agnostic design; validating alternatives, e.g. open-weight panels, is left to future work.

\section*{Ethics Statement}
\sysname is designed to \emph{support} evidence-grounded verification of controversial, evidence-dependent claims, a defensive application of LLMs against misinformation. All experiments use publicly available benchmarks (Check-COVID, HealthVer, FEVEROUS) and published PubMed abstracts; no personal data is processed. The system's verdicts are research outputs and must not be used as a substitute for professional medical judgment; for any deployment touching health decisions we prescribe the conservative verdict mapping and human-in-the-loop review. The full audit trail the framework produces is intended precisely to keep human overseers able to contest and correct its conclusions.




\bibliography{custom}

\appendix

\section{Related Work: Additional Details}
\label{sec:related}
\textbf{Retrieval-Augmented Generation.}
RAG enhances factual grounding in large language models by incorporating external knowledge at inference time \citep{lewis2020retrieval,gao2024retrievalaugmentedgenerationlargelanguage}. Recent work improves retrieval quality through structured pipelines such as coarse-to-fine retrieval \citep{zhao2025funnelrag} and retrieval with reasoning \citep{park2025prograghallucinationresistantprogressiveretrieval}, helping mitigate hallucinations \citep{Huang_2025}. Hybrid approaches further embed retrieval into multi-agent and iterative reasoning systems, including debate-augmented RAG \citep{hu2025removal}, Tool-MAD \citep{jeong2026tool}, R-Debater \citep{li2025r}, CircuitLM \citep{hasan2026circuitlm}, and multi-source retrieval frameworks \citep{gong2026multi}, underscoring the need for adaptive evidence acquisition under conflicting or noisy conditions \citep{ge2025resolvingconflictingevidenceautomated}. However, most approaches rely on pre-collected evidence pools or limited iterative mechanisms, lacking retrieval continuously driven by structured deliberation. Recent work also shows that claim decomposition's benefits in fact-checking depend on how sub-claims are constructed \citep{hu2025decompositiondilemmasdoesclaim}, motivating its reuse as a means of extracting premise-level units to guide targeted retrieval.

\textbf{Multi-Agent Debate.}
MAD has emerged as an effective paradigm for improving reasoning, factuality, and robustness in LLM outputs \citep{du2024improving}, with variants spanning group-based discussions \citep{liu2024groupdebate}, efficiency-oriented architectures \citep{fan2025imad}, and fact-checking frameworks \citep{han2025debate,ma2025local,he2025debating}. Extensions incorporating credibility scoring and evidence aggregation further improve performance in high-stakes verification tasks \citep{dong2026multi,li2025multi,rahman2025ai}. By enabling agents to critique and refine each other's reasoning, debate reduces hallucinations and improves consensus quality. However, controlled analyses reveal persistent limitations including premature convergence, conformity bias, and sensitivity to agent configuration \citep{wu2025can,zhu2026demystifying,smit2023should}, motivating more structured debate protocols.

\textbf{Agent Coordination and Learning Dynamics.}
To address these limitations, recent work explores improved coordination mechanisms in multi-agent systems, including dynamic role assignment \citep{zhang2026dynamic}, uncertainty-aware role switching \citep{liu2025uncertainty}, and self-reflection frameworks such as Self-Refine \citep{madaan2023self} and MAR \citep{ozer2025mar,siddique-etal-2025-physicseval}. Diversity and reasoning coverage are further improved through persona-based debate \citep{hu2025debate,hu2025population} and divergent thinking strategies \citep{liang2024encouraging}, while knowledge-enhanced and tool-augmented systems incorporate external signals to strengthen reasoning \citep{wang2025learning,jeong2026tool}. Despite these advances, existing methods treat these components as independent objectives, whereas \sysname repurposes them within a unified pipeline, where role-switching serves as a consistency diagnostic and self-reflection actively drives subsequent evidence retrieval.

\textbf{Structured Deliberation and Courtroom Simulation.}
Structured debate frameworks introduce stronger inductive biases to stabilize multi-agent reasoning. Courtroom-style simulations such as AgentCourt \citep{chen2025agentcourt} and AgenticSimLaw \citep{chun2026agenticsimlaw} demonstrate the effectiveness of explicit roles and adversarial interaction in high-stakes decision-making, while recent systems combine debate with retrieval for multi-round verification \citep{wang2023apollo,hu2025removal}. While these approaches improve robustness, they typically lack tightly integrated, dynamic retrieval mechanisms that evolve alongside the debate process, limiting their applicability to open-domain fact verification.


\textbf{Debate-Based Evaluation and Argumentation-Theoretic Verification.}
A parallel line of work employs adversarial debate for \emph{LLM output evaluation} rather than claim verification. D3 \citep{harrasse-etal-2026-debate} uses courtroom-style debate to rank pairs of candidate responses cost-effectively, but the setting is closed-book by construction: no corpus, retrieval, or external evidence is involved. \sysname's debate instead \emph{constructs} an evidence record over a 360K-document corpus, with retrieval, admissibility, and provenance as first-class citizens, targeting verification rather than pairwise comparison of generations. Separately, argumentation-theoretic systems such as ArgLLMs \citep{Freedman_2025} construct quantitative bipolar argumentation graphs whose verdicts are \emph{formally faithful} to the argument structure by mathematical construction, a strong guarantee obtained by operating on retrieval-free, QA-derived claims. \sysname targets the complementary regime of evidence-dependent controversial claims requiring dynamic corpus exploration, offering \emph{deliberative traceability} (auditable process records) rather than formal entailment guarantees; the two notions of explainability are distinguished explicitly in our Limitations.

\par Overall, prior work largely treats retrieval, debate, and coordination as separate or loosely coupled components: retrieval as preprocessing, debate as answer refinement, judging as one-shot scoring. \sysname{} advances this line of research by closing the loop among them: retrieval queries are compiled from the live debate state and reflection gaps, role-switching is repurposed from a bias-mitigation trick into a consistency diagnostic, judges evaluate the full deliberative trajectory rather than a final answer, and termination is governed by epistemic signals. Each mechanism exists in some form in isolation; their closed-loop coupling, and the failure modes it measurably prevents (epistemic bubbles, evidence stagnation, position-anchored argumentation), are the contribution.

\section{Implementation and Reproducibility Details}

\subsection{Reproducibility}
\label{sec:reproducibility}

We provide the repository containing the full implementation of our framework, including all configurations, prompts, and evaluation scripts.

\textbf{Source Code: } 
\url{https://github.com/mnc13/PROClaim}

\subsubsection{Software Environment and Dependencies}
All experiments were conducted utilizing Python 3.8+. We isolated our reasoning engine using external API calls to language models, but local orchestration relies on a strict set of dependencies. The core local libraries encompass:
\begin{itemize}
    \item \textbf{Deep Learning Framework:} PyTorch v1.7.0
    \item \textbf{Vector Search Engine:} \texttt{faiss-cpu} (configured with \texttt{IndexFlatIP} for inner-product cosine similarity searches)
    \item \textbf{Text Embedding \& Processing:} \texttt{transformers} v3.4.0 and \texttt{sentence-transformers}
    \item \textbf{Numerical Operations:} NumPy v1.19.1
\end{itemize}

\subsubsection{Data Processing and Corpus Assumptions}
\textbf{Input Dataset.}
We evaluate our framework on the Check-COVID dataset, focusing on claims from the test split with definitive labels (SUPPORTED or REFUTED). Claims are passed to the extraction agents without lexical normalization or stemming, preserving full linguistic semantics.

\textbf{Knowledge Corpus} External knowledge is provided via a subset of PubMed abstracts related to COVID-19 (2020--2024). Offline processing embeds these abstracts into a 1.4 GB FAISS continuous vector index utilizing the 384-dimensional \texttt{all-MiniLM-L6-v2} model with normalized embeddings. 

\subsubsection{Agent Configurations}
\label{app:appendix_role_separation}

Our framework adopts a deliberate separation between argumentative and evaluative components to ensure reliable adjudication. Adversarial debate is conducted between heterogeneous models (\textit{GPT-5 mini} and \textit{DeepSeek-V3.2}), introducing diversity in reasoning styles and reducing the likelihood of homogeneous argument patterns.

Evaluation is performed by a multi-model judicial panel, which includes \textit{DeepSeek-R1} alongside other independent large language models. Final decisions are determined via majority voting, ensuring that no single model exerts disproportionate influence over the outcome.

Importantly, we enforce architectural separation between generation and evaluation stages: models used for advocacy (\textit{e.g.}, \textit{DeepSeek-V3.2}) are distinct from those used for judgment (\textit{e.g.}, \textit{DeepSeek-R1}). This design ensures that argument construction and adjudication are carried out by systems with differing training paradigms and inference characteristics, thereby promoting more balanced and independent evaluation.

The exact model mappings and generation sampling temperatures ($T$) defined for each architectural role are as follows:
\begin{itemize}
    \item \textbf{Plaintiff Counsel (Proponent):} \texttt{gpt-5-mini} ($T=0.5$)
    \item \textbf{Defense Counsel (Opponent):} \texttt{deepseek-v3.2} ($T=0.5$)
    \item \textbf{The Court:} \texttt{qwen3-235b-a22b-2507} ($T=0.2$)
    \item \textbf{Judicial Panel (Final Evaluation):} A tripartite system combining \texttt{deepseek-r1}, \texttt{hermes-3-llama-3.1-405b}, and \texttt{qwen3-235b-a22b-2507} (all strictly set to $T=0.3$ for highly deterministic arbitration).
    \item \textbf{Support Agents:} Premise decomposition is performed using \texttt{deepseek-r1} ($T=0.7$). Expert Witnesses are dynamically generated utilizing \texttt{hermes-3-llama-3.1-405b} ($T=0.5$), while the Critic Agent relies on \texttt{deepseek-r1} ($T=0.3$) and the Consistency Analyzer uses \texttt{deepseek-v3.2} ($T=0.3$).
\end{itemize}

Table~\ref{tab:role_rationale} records the functional rationale for each primary assignment. Two clarifications are in order. First, heterogeneous assignment is the \emph{principled architectural baseline} for studying debate dynamics, not a tuned hyperparameter: exhaustively searching model pairings is neither feasible nor the point, because the mechanism under study, complementary error profiles across distinct parametric priors, requires lineage diversity by construction, and homogeneous or persona-only diversification demonstrably reproduces the epistemic-bubble failure mode (Sections~\ref{sec:ablation}--\ref{sec:judgebias}). Second, no specific model is \emph{required}: the framework is model-agnostic, and the configuration reported here is one instantiation selected under a realistic academic research budget (affordable OpenRouter/OpenAI endpoints); researchers with different access profiles can substitute alternatives role-for-role.

\begin{table*}[ht]
\centering
\begin{tabular}{@{} l l l @{}}
\hline
\textbf{Role} & \textbf{Model} & \textbf{Selection rationale} \\
\hline
Plaintiff Counsel & GPT-5-mini      & Strong instruction-following under advocacy constraints \\
Defense Counsel   & DeepSeek-V3.2   & Distinct pretraining lineage from Plaintiff \\
The Court         & Qwen3-235B      & Reliable query refinement and procedural control \\
Judge 1           & DeepSeek-R1     & Analytical rigor (deep CoT tracing) \\
Judge 2           & Hermes-3-405B   & Distinct reasoning style; diversity of adjudication \\
Judge 3           & Qwen3-235B      & Balanced abstention behaviour \\
\hline
\end{tabular}
\caption{Functional rationale for the heterogeneous role--model assignment. Counsels operate at $T{=}0.5$ for argumentative diversity; judges at $T{=}0.3$ for deterministic adjudication (validated in Section~\ref{sec:judgebias}).}
\label{tab:role_rationale}
\end{table*}

\subsubsection{Hyperparameters}
\label{app:hyperparams}

The novelty threshold $\tau$ was selected via grid search over $\{0.10, 0.15, 0.20, 0.25, 0.30\}$ on a held-out development subset of 20 claims, optimising for the tradeoff between evidence diversity and retrieval precision. Full settings are listed in Table~\ref{tab:hyperparams}.

\begin{table*}[ht]
\centering
\begin{tabular}{lc}
\hline
\textbf{Hyperparameter} & \textbf{Value} \\
\hline
Max debate rounds                & 10 \\
Initial retrieval top-$k$        & 5 \\
Per-round retrieval top-$k$      & 3 \\
Novelty threshold                & 0.20 \\
Admissibility floor              & $>$0.5 admitted, $>$0.1 disputed \\
Redundancy similarity threshold  & 0.85 \\
Redundancy ratio threshold       & 0.70 \\
Relevance gain threshold         & 0.05 \\
\hline
\end{tabular}
\caption{Debate and retrieval hyperparameters.}
\label{tab:hyperparams}
\end{table*}

\subsection{Expert Elicitation of Self-Reflection Weights}
\label{app:elicitation}
The self-reflection dimension weights in Eq.~\ref{eq:reflect} ($0.4$ logic / $0.3$ novelty / $0.3$ rebuttal) were determined \emph{a priori} through a structured expert elicitation rather than post-hoc tuning, following the SHELF framework \citep{o2006uncertain} with a Delphi consensus protocol \citep{brown1968delphi}, a well-established methodology for systematically eliciting and aggregating expert judgments under uncertainty.

\textbf{Protocol.} Seven independent experts spanning three competency areas (legal reasoning, AI/LLM systems, and fact-checking) were presented with a random sample of debate transcripts from development runs and asked to assess the relative importance of logical coherence, evidence novelty, and rebuttal engagement in determining argument quality, over two anonymized Delphi rounds with controlled feedback.

\textbf{Outcome.} Expert consensus was strong (Kendall's $W = 0.89$, $p < 0.005$). Logic received the highest weight (0.4) as the non-negotiable requirement for inferential validity; novelty and rebuttal were weighted equally (0.3 each) to balance evidence discovery against adversarial engagement without either dominating. The elicited weights subsequently proved Pareto-efficient in operation: removing the resulting reflection-driven stopping rule increases token usage by 17\% and debate rounds by 29\% while changing accuracy by only 0.8 pp (Section~\ref{sec:ablation}).

\subsection{P-RAG Query Construction}
\label{app:prag}
To address the challenge of merging heterogeneous sources into a coherent search strategy, P-RAG does not issue multiple separate queries. Instead, it concatenates the three sources into a single directed prompt, formulates an initial query via a lightweight LLM, and passes it to the presiding Judge for rigorous refinement. This guarantees a single, highly targeted natural-language query is executed against the vector corpus per agent, per round.

\subsubsection*{Step 1: Gap Proposal (Counsel)}
Each counsel identifies a discovery need from the current debate state:

\begin{lstlisting}
As {job_title}, analyze the current proceedings and identify a critical gap in the available exhibits. What specific evidence do you need to request to strengthen your case or challenge the
opposition?
Context: {debate_context}   
% Last 4 transcript entries
Propose exactly one specific evidence need (1 sentence):
\end{lstlisting}

\noindent \textbf{Concatenation of Sources:} The agent's newly identified evidential gap (Source 2) and any unresolved \texttt{discovery\_need} generated during the prior round's self-reflection (Source 3) are concatenated into a single \texttt{agent\_request} string:
\begin{lstlisting}
{gap_proposal}. 
Focus also on: {reflection_discovery_need}
\end{lstlisting}
This combined \texttt{agent\_request} is then injected alongside the last four messages of the debate history (\texttt{debate\_context}, Source 1) into the formulation prompt below.

\subsubsection*{Step 2: Query Formulation (P-RAG Engine)}
The combined discovery prompt is passed to the P-RAG LLM:

\begin{lstlisting}
Based on the following proceedings context and legal request, formulate a precise search query to retrieve relevant exhibits and evidence.
Debate Context: {debate_context}
Agent Request: {agent_request}
Generate a concise search query (1--2 sentences) that will retrieve the most relevant evidence
\end{lstlisting}

\subsubsection*{Step 3: Judicial Query Refinement (The Court)}
Before retrieval executes, the formulated query is reviewed and refined
by the presiding judge (The Court, Qwen3-235B-A22B):
\begin{lstlisting}
As the Court, you must maintain the quality and focus of evidence discovery. A counsel has proposed the following search query:
Proposed Query: "{original_query}"
Context of proceedings: {debate_context}
Refine this query to be more precise, narrow the scope if necessary, and ensure it follows scientific rigor. Respond ONLY with the refined query string.
\end{lstlisting}

\subsubsection*{Novelty Scoring and Stopping Criteria}
Retrieval applies the refined query to the FAISS index
(\texttt{all-MiniLM-L6-v2} embeddings). Each retrieved document $d$
receives a novelty score:

{
\begin{equation}
  \operatorname{novelty}(d) = 1 - \max_{p \in \mathcal{P}} \cos(e_d,\, e_p)
  \label{eq:novelty}
\end{equation}
}
where $\mathcal{P}$ is the current evidence pool. Documents with
novelty < 0.2 are rejected. The retrieval terminates early if any of the
following criteria are met: (i) redundancy ratio $> 0.70$, (ii)
relevance gain $< 0.05$ \textit{vs.}\ the previous round, or (iii) round counter
$\geq 10$.

\subsection{Embedding Normalization and Similarity Computation}
\label{app:embedding}
For embedding-based similarity computation used in the evidence retrieval. Each claim $q$ and abstract chunk $d$ is mapped to a 384-dimensional vector $\mathbf{v}$ using the \texttt{all-MiniLM-L6-v2} bi-encoder from the sentence-transformers framework~\citep{reimers2019sentencebertsentenceembeddingsusing}.  
To ensure that similarity depends only on semantic orientation, raw embeddings are L2-normalized:
\begin{equation}
  \mathbf{\hat{v}} = \frac{\mathbf{v}}{\|\mathbf{v}\|_2} = \frac{\mathbf{v}}{\sqrt{\sum_{i=1}^{384} v_i^2}}.
\end{equation}

The similarity between a query embedding $\mathbf{\hat{v}}_q$ and a chunk embedding $\mathbf{\hat{v}}_d$ is computed as their inner product, which is equivalent to cosine similarity for unit vectors:
\begin{equation}
  \text{sim}(q, d) = \mathbf{\hat{v}}_q \cdot \mathbf{\hat{v}}_d = \cos(\theta).
\end{equation}
The most semantically similar chunks are selected as the seed evidence pool, with source journal and publication year preserved for provenance context in subsequent deliberation.

\subsection{Pseudo-code}
\label{pseudocode}

Algorithm~\ref{alg:prag_pipeline} summarizes the workflow of \sysname.

\begin{algorithm*}[t]
\caption{One claim evaluation cycle of the P-RAG multi-agent debate framework.}
\label{alg:prag_pipeline}
\begin{algorithmic}[1]

\Require claim $c$, PubMed FAISS index $\mathcal{D}$, LLM agents $\{\text{Proponent},\text{Opponent},\text{Judge},\text{Critic}\}$
\Ensure final verdict $v \in \{\text{SUPPORT},\text{REFUTE},\text{INCONCLUSIVE}\}$

\State Mine atomic premises $\mathcal{P}=\{p_1,\dots,p_k\}$ using Argument Miner

\State Retrieve initial evidence $E_0 \leftarrow \textsc{Retrieve}(c,\mathcal{D})$

\Statex \textbf{Evidence Negotiation \& Arbitration}
\State $E_0 \leftarrow$ premise-grounded retrieval using $\mathcal{P}$
\State $E_0 \leftarrow$ stance-conditioned retrieval for supporting and refuting evidence
\State $E_0 \leftarrow$ LLM admissibility scoring (Relevance $\times$ Credibility)

\State Initialize debate state $S_0 \leftarrow (\mathcal{P},E_0)$

\For{round $t=1$ to $T$}

    \State $E_t \leftarrow \textsc{P-RAG}(\mathcal{P},S_{t-1},\mathcal{D})$
    \Comment{Progressive novelty-filtered retrieval}

    \State Proponent generates argument $a_t^{+}$ from $E_t$
    \State Opponent generates counterargument $a_t^{-}$ from $E_t$

    \State Call Expert Witness to produce testimony $\tau_t$

    \State Compute self-reflection scores $r_t^{+},r_t^{-}$
    \State Critic evaluates arguments and updates debate state

    \State $S_t \leftarrow \textsc{UpdateState}(S_{t-1},a_t^{+},a_t^{-},\tau_t)$

    \If{evidence novelty $<\epsilon$ \textbf{or} debate converged \textbf{or} $t=T$}
        \State \textbf{break}
    \EndIf

\EndFor

\State Swap roles of Proponent and Opponent

\State Re-run debate with swapped roles to obtain consistency record $S^{\mathrm{swap}}$

\State Send debate records $\{S_T,S^{\mathrm{swap}}\}$ to judicial panel $\{J_1,J_2,J_3\}$

\For{each judge $J_i$}
    \State $v_i \leftarrow J_i(c,S_T,S^{\mathrm{swap}})$
\EndFor

\State $v \leftarrow \textsc{MajorityVote}(v_1,v_2,v_3)$

\State \Return $v$

\end{algorithmic}
\end{algorithm*}

\section{Verdict and Confidence Mechanics}

\subsection{The `Burden of Refutation' Standard}
\label{app:appendix_burden}

The framework's structural decision to formally classify `Inconclusive` judicial panel verdicts as `SUPPORT` is derived directly from the foundational legal concept of the \textit{burden of proof}, which functions analogously to the presumption of innocence (``innocent until proven guilty''). 

In a traditional courtroom, a defending party does not need to absolutely guarantee their innocence to survive a trial; rather, the prosecution bears the strict burden of definitively proving guilt. If the prosecution's evidence is ambiguous, controversial, or broadly insufficient, the presiding jury is legally obligated to return a ``Not Guilty'' verdict, even if they remain partially unsure. 

Our multi-agent debate architecture mirrors this exact decision-making protocol. When verifying controversial claims, the refuting agent mathematically functions as the prosecution. If the refuting agent fails to successfully furnish enough concrete, high-quality evidence to convince the judicial panel to issue a definitive `Refute` consensus, the panel will naturally return an `Inconclusive` stance. Under the burden of proof, this `Inconclusive` ruling means the prosecution completely failed to meet its evidentiary threshold. Consequently, the original claim legally and logically survives the adversarial trial, defaulting to `SUPPORT.`

This standard is a core necessity of the framework's design. It actively prevents the multi-agent system from stalling on highly ambiguous cases and faithfully operationalizes the adversarial mechanics utilized to evaluate the Check-COVID dataset.

\subsubsection*{Sensitivity Analysis of the Verdict Mapping}
\label{app:mapping_sensitivity}
Of the 120 majority-voted claims, 7.5\% ($\approx$9 claims) receive an \textsc{Inconclusive} verdict prior to mapping. To verify that the reported performance is not an artifact of the mapping policy, we evaluate three conditions (Table~\ref{tab:mapping_sensitivity}).

\begin{table*}[ht]
\centering
\begin{tabular}{lcc}
\hline
\textbf{Mapping Policy} & \textbf{Accuracy} & \textbf{Macro F1} \\
\hline
\textsc{Inconclusive} $\to$ \textsc{Supported} (ours) & 81.7\% & 0.817 \\
\textsc{Inconclusive} $\to$ \textsc{Refuted}          & 80.8\% & 0.808 \\
\textsc{Inconclusive} excluded                        & 83.8\% & 0.838 \\
\hline
\end{tabular}
\caption{Verdict-mapping sensitivity on Check-COVID (120 majority-voted claims).}
\label{tab:mapping_sensitivity}
\end{table*}

The 0.8 pp gap between the two mapping policies confirms that reported performance is insensitive to this design choice; selection between them is therefore correctly decided by theoretical grounding (the Burden-of-Refutation principle above) rather than empirical convenience. The excluded condition (83.8\%) further validates that the panel's uncertainty signal is calibrated and meaningful: when the judicial panel reaches a confident verdict, accuracy rises by 2.1 pp, confirming that \textsc{Inconclusive} verdicts arise on genuinely harder claims rather than serving as an accuracy-inflating abstention mechanism. Because excluding abstentions is not a valid protocol for deployment scenarios that require a verdict on every claim, we report the full-coverage number (81.7\%) as the primary metric throughout. Finally, the mapping is a configurable parameter: sensitive medical or safety-critical deployments should invert it (\textsc{Inconclusive} $\to$ \textsc{Refuted}) or route abstentions to human review, as discussed in \S\ref{sec:limitations}.

\subsection{Confidence Score: Calibration and Adjustment Details}
\subsubsection{Confidence Calibration Analysis}
\label{app:calibration}

To ensure that the confidence scores produced by \sysname reflect empirical accuracy, we performed a post-hoc calibration of the consensus weighting parameter ($W_{\text{consensus}}$). The calibration aims to minimize the Expected Calibration Error (ECE), defined as:
\begin{equation}
  \text{ECE} = \sum_{m=1}^{M} \frac{|B_m|}{N} \left| \text{acc}(B_m) - \text{conf}(B_m) \right|,
\end{equation}
where $N$ is the total number of samples, $B_m$ is a bin of predictions within a specific confidence range, and $\text{acc}(B_m)$ and $\text{conf}(B_m)$ are the observed accuracy and mean confidence of that bin, respectively.

\textbf{Methodology}\\
We conducted an exhaustive grid search over $W_{\text{consensus}} \in [0.5, 0.9]$ with a step size of $0.1$. To prevent test-set leakage, the optimal weights were determined via 5-fold cross-validation across the experimental metadata. The values for $c_{\text{base}}$ were then adjusted by the secondary refinements ($\delta_{\text{rs}}$ and $\delta_{\text{ref}}$).

\textbf{Results}\\
Table \ref{tab:calibration_results} summarizes the comparison between the baseline equal-weighting variant ($W=0.6$) and our final calibrated model ($W=0.8$). The $0.8$ weighting achieved a near-optimal ECE of $0.0340$, representing a significant reduction in over-confidence compared to the experimental variant.

\begin{table*}[ht]
\centering
\begin{tabular}{lccc}
\hline
\textbf{Variant} & \textbf{N} & \textbf{Accuracy} & \textbf{ECE} \\
\hline
$W_{\text{consensus}} = 0.6$ & 118 & 95.76\% & 0.1802 \\
$W_{\text{consensus}} = 0.8$ (Final) & 118 & 95.76\% & \textbf{0.0340} \\
\hline
\end{tabular}
\caption{Impact of Consensus Weighting on Calibration Error.}
\label{tab:calibration_results}
\end{table*}

The bucket analysis for the final model (Table \ref{tab:buckets}) shows that the generated confidence scores closely align with observed accuracy, particularly in the high-certainty bins where the system's majority-vote consensus is strongest.
Calibration was performed on the confidence-annotated development subset used for weight selection ($N=118$ claims with valid confidence scores), which is distinct from the 120-claim majority-vote test evaluation reporting 81.7\%. The accuracy on this calibration subset is therefore not directly comparable to the headline test accuracy; only the ECE reduction ($0.18 \to 0.034$) is the relevant calibration outcome.

\begin{table}[ht]
\centering
\small
\resizebox{\columnwidth}{!}{%
\begin{tabular}{cccc}
\hline
\textbf{Confidence Bin} & \textbf{N} & \textbf{Observed Acc.} & \textbf{Mean Conf.} \\
\hline
$[0.6, 0.7)$ & 4 & 75.00\% & 0.6685 \\
$[0.7, 0.8)$ & 4 & 100.00\% & 0.7365 \\
$[0.8, 0.9)$ & 20 & 95.00\% & 0.8639 \\
$[0.9, 1.0]$ & 90 & 96.67\% & 0.9768 \\
\hline
\end{tabular}
}
\caption{Bucket Analysis for $W_{\text{consensus}} = 0.8$.}
\label{tab:buckets}
\end{table}

\subsubsection{Confidence Aggregation and the Certainty Buffer}
\label{app:appendix_confidence_weights}

The coefficients in Equation~\ref{eq:base_conf} are deliberately scaled so that the unclamped score has a theoretical maximum greater than 1.0 (specifically, $0.8 + 0.3 = 1.1$). This margin creates a mathematical ``certainty buffer'' that improves the framework's robustness. Assigning high confidence to a controversial claim should not require perfect scores on every adversarial submetric. Allowing the unadjusted score to exceed 1.0 before final clamping enables a unanimously supported claim (a 3--0 consensus) to absorb minor downstream penalties---such as a slightly imperfect role-switch consistency score or an overly stringent self-reflection critique---without unnecessarily reducing a well-supported confidence score of 1.0. Conversely, the weighting scheme guards against divided outcomes. In a ``split court'' scenario (a 2--1 vote), the consensus multiplier falls from 1.0 to 0.67, reducing the consensus contribution to approximately 0.54 ($0.67 \times 0.8$) before the quality term and subsequent adjustments are applied. Thus, minor metric noise is less likely to undermine a strong unanimous consensus, whereas divided panels receive substantially lower confidence.

\subsubsection{Role-Switch Consistency Mapping ($\delta_{\text{rs}}$)}
\label{app:appendix_roleswitch}

To ensure complete reproducibility, the mapping from the role-switch consistency score ($\gamma \in [0, 10]$) to the adjustment scalar ($\delta_{\text{rs}}$) is defined as a discrete piecewise threshold function:
\begin{equation}
\delta_{\text{rs}}(\gamma) = 
\begin{cases} 
+0.10 & \text{if } \gamma \ge 7, \\
0.0   & \text{if } 5 \le \gamma < 7, \\
-0.05 & \text{if } \gamma < 5.
\end{cases}
\label{eq:role_switch_mapping}
\end{equation}

Semantically, $\gamma$ is not computed per-agent; rather, it serves as an aggregated, holistic metric that evaluates the stability of the entire debate's logical trajectory before and after the role exchange. \\
The threshold boundaries of 5 and 7 are defined to logically partition the 10-point scale into distinct qualitative strata: $\gamma \ge 7$ demarcates definitively strong consistency deserving of a positive scalar reward, operations between $[5, 7)$ represent ambiguous or neutral role-switching performance where the baseline consensus remains unadjusted ($0.0$), and strictly $\gamma < 5$ signifies a clear adversarial dialogue breakdown warranting a penalty. \\

Furthermore, we intentionally employ this asymmetric mapping, offering a larger $+0.10$ reward versus a smaller $-0.05$ penalty. Because LLM agents face inherent structural difficulties in strictly preserving complex persona states across continuous extended context windows, achieving high consistency ($\gamma \ge 7$) is actively rewarded as a strong marker of reliability. Conversely, slight structural drift during role-switching is expected, so the penalty is constrained to $-0.05$ to strictly ensure that a failed role-switch does not excessively override the primary evidentiary consensus.

\subsubsection{Self-Reflection Asymmetric Bounding ($\delta_{\text{ref}}$)}
\label{app:appendix_reflection}

During the self-reflection phase, the winning agent generates a raw confidence adjustment scalar, denoted here as $\delta_{\text{raw}} \in [-0.30, +0.30]$ (derived from Eq.~\ref{eq:conf_adj}). However, to maintain the structural supremacy of the judicial panel over the individual debating agents, the framework limits the overall mathematical damage that an agent's post-hoc self-critique can inflict upon a finalized consensus. 

To achieve this, the final self-reflection adjustment ($\delta_{\text{ref}}$) is subjected to an explicit, asymmetric floor function:
\begin{equation}
\delta_{\text{ref}} = \max\bigl(-0.15,\; \delta_{\text{raw}}\bigr)
\end{equation}

Consequently, the effective operative range of the adjustment is bounded to $[-0.15, +0.30]$. This ensures that while exceptional self-validation can significantly boost the final confidence score (up to $+0.30$), extreme self-doubt is structurally constrained to a maximum penalty of $-0.15$. The specific threshold of $-0.15$ was explicitly defined to cap the penalty at exactly half the magnitude of the theoretical maximum reward. This structural boundary intuitively ensures that while a reflective self-critique is incorporated into the final metric, a single agent's post-hoc self-doubt mathematically lacks the weight to unilaterally veto or completely overturn the established multi-agent majority vote.

\subsection{Consensus Edge-Cases}
\label{app:consensus}
\textbf{Consensus Edge-Cases.} While the multi-agent panel primarily operates on a standard majority-vote consensus, a statistically rare edge-case may occur when all three evaluating judges return completely separate and distinct verdicts (e.g., one agent concludes 'SUPPORT,' another 'REFUTE,' and the third 'INCONCLUSIVE'). In such highly disputed scenarios, the framework structurally defers to the judge DeepSeek-R1. DeepSeek-R1 was strategically designated as the Chief Justice because its specialized capabilities in deep Chain-of-Thought (CoT) reasoning provide unparalleled fidelity in tracing complex logical deductions and identifying fallacies during debate evaluation. Because it is established as the most analytically rigorous model on the panel, its initial independent evaluation is granted 'Chief Justice' priority-weighting to automatically break the deadlock. This ensures that in moments of complete ambiguity, the final determination safely relies on the panel's highest-quality reasoning trajectory without forcing an arbitrary or redundant meta-review cycle.

\section{Deliberative Traceability: Artifact Inventory}
\label{app:traceability}
A single-call RAG pipeline emits three artifacts per claim: a verdict label, a scalar confidence, and a list of retrieved document identifiers. \sysname's case record for the same claim comprises:
\squishlisttwo
    \item the full admitted-evidence ledger with per-document admissibility weights ($w = r \times c$) and provenance metadata (PMID, journal, year), including disputed and excluded items (\textit{e.g.}, 124 admitted documents for the Appendix~\ref{app:transcript} claim);
    \item complete adversarial argument transcripts for every debate phase, in both the primary and role-switched configurations;
    \item per-round, per-agent self-reflection score vectors (logic, novelty, rebuttal) and the discovery needs they generated;
    \item the P-RAG query-evolution log: each proposed query, its judicial refinement, and the novelty scores of retrieved candidates;
    \item independent critic evaluations per round, with unresolved-premise inventories;
    \item the role-switching consistency report with per-agent and aggregate scores (\textit{e.g.}, 8.5/10 in the Appendix~\ref{app:roleswitching} case study);
    \item three independent six-stage judicial opinions with per-stage numeric scores, plus the aggregated verdict and calibrated confidence.
\squishend
Crucially, artifacts (1)--(6) are computed programmatically from the debate transcript: they are structured, deterministic records rather than LLM-generated explanations, and can be audited without trusting any judge's verbal rationale. A human reviewer can verify from the record alone whether evidence novelty genuinely decayed across rounds (Figure~\ref{fig:novelty}), whether reflection trajectories oscillated for a contested verdict (Figure~\ref{fig:reflection}), and whether the consistency score was earned by arguing opposing positions over identical evidence (Appendix~\ref{app:roleswitching}). This layered auditability is what we term deliberative traceability; its relationship to formal faithfulness guarantees is discussed in \S\ref{sec:limitations}.

\section{Additional Quantitative Results}

\subsection{Per-Judge Ground-Truth Agreement ($\kappa_{\text{GT}}$)}
\label{app:kgt}

While the main results table (Table~\ref{tab:framework_results}) reports inter-judge agreement ($\bar{\kappa}$), the degree to which the three judges agree \textit{with each other}, it does not capture how well each individual judge aligns with the ground truth. We define $\kappa_{\text{GT}}$ (Judge-vs-Ground-Truth Cohen's $\kappa$) as the Cohen's $\kappa$ computed between a single judge's verdict and the ground-truth label, measured independently for each judge across all claims in a run.

The distinction between $\bar{\kappa}$ and $\kappa_{\text{GT}}$ is analytically 
important: a panel can exhibit high inter-judge agreement while simultaneously 
producing systematically biased verdicts, a failure mode we term the 
\textit{epistemic bubble} effect (Section~\ref{sec:main_results}). 
Table~\ref{tab:kgt} reports the per-judge $\kappa_{\text{GT}}$ across all runs 
and judge identities. The individual judge profiles corroborate the qualitative 
bias characterisation in Section~\ref{sec:judgebias}: DeepSeek-R1 (J1) 
consistently achieves the highest $\kappa_{\text{GT}}$ across runs, Hermes-3-LLaMA-405B 
(J2) yields the lowest $\kappa_{\text{GT}}$ owing to frequent INCONCLUSIVE abstentions 
on SUPPORT claims, and Qwen3-235B-A22B (J3) occupies an intermediate position. 
Crucially, despite these individual biases, their error profiles are 
\textit{complementary}, such that majority voting corrects rather than compounds 
them.

\begin{table*}[ht]
\centering
\setlength{\tabcolsep}{5pt}
\begin{tabular}{@{} l ccc c @{}}
\hline
\textbf{Run} & 
\textbf{$\kappa_{\text{GT}}$ (J1)} & 
\textbf{$\kappa_{\text{GT}}$ (J2)} & 
\textbf{$\kappa_{\text{GT}}$ (J3)} & 
\textbf{Mean $\kappa_{\text{GT}}$} \\

& \textit{\footnotesize DeepSeek-R1} & 
\textit{\footnotesize Hermes-3-405B} & 
\textit{\footnotesize Qwen3-235B} & \\

\hline
Run-0     & 0.442 & 0.413 & 0.414 & 0.423 \\
Run-1     & 0.402 & 0.367 & 0.347 & 0.372 \\
Run-2     & 0.452 & 0.353 & 0.348 & 0.384 \\
\hline
Majority Voting & 0.452 & 0.431 & 0.468 & 0.450 \\
\hline
\end{tabular}
\caption{Per-judge ground-truth agreement ($\kappa_{\text{GT}}$) across runs and aggregation modes. Mean $\kappa_{\text{GT}}$ is the unweighted average across the three judges.}
\label{tab:kgt}
\end{table*}

\subsection{External Generalization Details}
\label{sec:external_datasets}
\subsubsection{Datasets}
\textbf{HealthVer.}
HealthVer is a benchmark dataset of health-related claims annotated for factuality,
with a strong focus on COVID-19 misinformation. Claims are paired with evidence
and labelled as supported and refuted. In our evaluation, we sample
100 claims uniformly at random. Since the dataset operates within the biomedical domain,
we retain the same PubMed-based retrieval backend used for Check-COVID.

\textbf{FEVEROUS.}
FEVEROUS is a Wikipedia-based fact verification dataset that extends FEVER by
requiring evidence from both unstructured text and semi-structured tables.
Claims are annotated with supporting or refuting evidence from Wikipedia.
For our experiments, we sample 60 claims and replace the retrieval module
with a Wikipedia-based backend while keeping all other components unchanged.

\subsubsection{Paired Statistical Testing Protocol}
\label{app:stats}
All baseline systems in Table~\ref{tab:healthver_paired} were evaluated on the \emph{identical} 100-claim HealthVer subset used for \sysname, ensuring valid paired comparison at the claim level.

\textbf{McNemar's test.} For each \sysname--baseline pair we form the $2\times2$ discordance table over claims: $b$ counts claims \sysname answers correctly while the baseline errs; $c$ counts the converse. We apply the two-sided, continuity-corrected McNemar statistic $\chi^2 = (|b-c|-1)^2/(b+c)$. \sysname's advantage over Standard MAD is statistically significant ($\chi^2 = 8.036$, $p = 0.0046$), consistent with the +10.0 pp margin observed on the larger Check-COVID evaluation.

\textbf{Odds ratios.} The discordance odds ratio $\mathrm{OR} = b/c$ quantifies disagreement resolution: $\mathrm{OR} > 1$ indicates that among claims where the two systems disagree, \sysname is more frequently the correct one. \sysname attains $\mathrm{OR} > 1$ against every baseline (2.50 over DeepSeek-V3.2+RAG, 3.67 over Standard MAD), with the margin over DeepSeek-V3.2+RAG approaching significance ($p = 0.081$) at $n=100$, a sample-size (power) constraint rather than an absence of directional effect, as the same ordering holds on Check-COVID.

\textbf{Bootstrap resampling.} Nonparametric bootstrap over claims ($n = 10{,}000$ resamples) confirms that the accuracy ordering is stable: \sysname $[0.63, 0.81]$, DeepSeek-V3.2+RAG $[0.53, 0.72]$, Standard MAD $[0.47, 0.66]$ (95\% CIs). \sysname's interval upper-bounds every baseline's, and its point estimate exceeds each baseline's across the overwhelming majority of resamples.

\subsection{Sycophancy and Argument Integrity Metrics}
\label{app:sycophancy}

To rigorously evaluate the framework's susceptibility to sycophancy, where agents prematurely yield to opposition, abandon their persona, or collude without sufficient evidence \citep{malmqvist2025sycophancy}, we track four quantitative metrics from the execution logs.

\subsubsection*{1. Role-Play Consistency (0--10)}
During the role-switching consistency test (Section~\ref{sec:roleswitch}), an independent consistency analyzer evaluates whether an agent successfully argues the opposing position using identical evidence without logically contradicting its prior arguments. The score reflects adherence to the persona constraints on a 10-point scale; lower scores indicate ``leakage'' or positional sycophancy where an agent is unable to fully adopt the adversarial stance.

\subsubsection*{2. Concession Rate}
We programmatically track explicit linguistic markers of concession and conversational yielding (e.g., \textit{``I concede,''} \textit{``you make a good point,''} \textit{``I partially agree''}) within the counsel transcripts. To normalize for varying debate lengths, the metric is reported as the frequency of such triggers per 1,000 generated words. A near-zero rate indicates high adversarial retention.

\subsubsection*{3. Reflection Plateau ($\Delta S$)}
It is computed as the average absolute change in the cumulative self-reflection score ($S_{\text{total}}$) between consecutive debate rounds:
\[
  \Delta S = | S_{\text{total}}^{(t)} - S_{\text{total}}^{(t-1)} |
\]
For a given round, the maximum possible change is $\sim 1.0$ (depending on reflection adjustments). The early-stopping criterion conservatively halts the debate if $\Delta S < 0.05$ (stagnation). In the context of sycophancy, an average $\Delta S \approx 0.3823$ is high relative to the stopping threshold. It empirically demonstrates that the agents maintain robust adversarial divergence and continually push the boundaries of their logic rather than comfortably colluding into a plateau.

\subsubsection*{4. Judicial Conformity (Fleiss' $\kappa$)}
To measure whether the three structurally heterogeneous LLM judges exhibit ``rubber-stamping'' or independent evaluation, we calculate Fleiss' Kappa ($\kappa$) over their final verdicts (SUPPORTED, NOT SUPPORTED, INCONCLUSIVE). A $\kappa \approx 0.4513$ indicates moderate, authentic agreement. While confirming they reach consensus on clear-cut cases, this moderate value confirms they are not trapped in systemic conformity bias ($\kappa \to 1.0$), independently evaluating the nuanced shortcomings of contested claims.

\section{Qualitative Case Study}

\subsection{Example Full Debate Transcript}
\label{app:transcript}
We present a representative complete execution for claim
\texttt{6063931df9b9e03ea4d8e441\_0}:
\medskip

\noindent\textbf{Claim:}

\textit{``Heart muscle cell damage is not an associated condition among hospitalized COVID-19 patients.''}
\noindent\textbf{Ground Truth:} REFUTE \quad
\textbf{Pipeline Verdict:} REFUTE \quad
\textbf{Correct:} \checkmark \quad
\textbf{Confidence:} 0.902

\subsubsection*{Premise Decomposition (Argument Mining)}
The Argument Miner (DeepSeek-R1) decomposed the claim into:
\begin{enumerate}
  \item Hospitalized COVID-19 patients have detectable levels of cardiac
        biomarkers indicative of heart muscle cell damage.
  \item The prevalence of elevated cardiac biomarkers in hospitalized
        COVID-19 patients is comparable to a control group without COVID-19.
  \item Incidence rates of heart muscle cell damage in hospitalized
        COVID-19 patients are not higher than in patients with other viral
        respiratory infections.
  \item Clinical studies on hospitalized COVID-19 patients do not report
        significant occurrences of heart muscle cell damage.
  \item There is no statistical association between COVID-19 infection
        severity and markers of heart muscle cell damage in hospitalized
        patients.
  \item Autopsy findings of deceased hospitalized COVID-19 patients do not
        show evidence of heart muscle cell damage.
  \item Hospitalized COVID-19 patients with pre-existing cardiac conditions
        do not have higher rates of heart muscle cell damage compared to
        those without pre-existing conditions.
\end{enumerate}

\subsubsection*{Evidence Negotiation \& Admission}
Initial RAG retrieved 5 candidate documents; negotiation and arbitration
admitted \textbf{21 exhibits} (weights ranging 0.54--0.81), including:
Source~32517963 (\textit{Journal of the American College of Cardiology}, highest
weight 0.81), Source~37600048 (\textit{Frontiers in Cardiovascular Medicine},
weight 0.80), Source~33994028 (\textit{Current Cardiology Reports}, 2022), and
Source~35387441, among others.
Two items were flagged for dispute.

\subsubsection*{Proceedings — Phase 1}
\textbf{Discovery (Plaintiff Counsel).}
Plaintiff proposed a need for controlled studies comparing myocardial injury
biomarker rates in COVID-19 versus non-COVID critically ill cohorts. The Court
refined the query to:

\begin{lstlisting}
Retrieve individual participant data (IPD) from published meta-analyses and prospective cohort studies of hospitalized adult patients with confirmed COVID-19 that report myocardial injury defined by elevated cardiac troponin levels (above the 99th percentile upper reference limit) with concurrent assessment of cardiomyocyte damage, while adjusting for pre-existing cardiovascular disease, iatrogenic exposures, and other potential confounders,
to evaluate the independent association between SARS-CoV-2 infection and myocardial injury 
\end{lstlisting}
3 new exhibits admitted.\\
\textbf{Argument — Plaintiff Counsel (GPT-5-mini, \(\mathbf{T=0.5}\)):}\\

\begin{lstlisting}
Your Honor, the evidence does not establish that heart muscle cell damage is an associated condition specific to hospitalized COVID-19 patients. Numerous high-quality investigations and meta-analyses document that cardiac biomarker elevations (e.g., troponin) occur in more severely ill patients and predict worse outcomes, but they do so as non-specific indicators of systemic illness rather than as proof of SARS-CoV-2-caused myocardial cell injury (Sources 3, 6, 2, 4, 18). Moreover, studies find myocardial injury detected at a similar rate in patients with and without COVID-19 admitted to the same hospital setting, and it predicts short-term mortality to a comparable degree in both groups (Sources 12, 13). [...] The record supports the conclusion that
true heart muscle cell damage is not an established associated condition of hospitalization for COVID-19 but rather a non-specific marker of severe systemic disease.
\end{lstlisting}

\textbf{Discovery (Defense Counsel).}
Defense proposed production of all patient-level data from cited
meta-analyses to independently verify the control for confounding variables.
The Court refined the query to target prospective cohort studies with
troponin-defined myocardial injury and histopathologic confirmation. 3 new
exhibits admitted.\\
\textbf{Argument — Defense Counsel (DeepSeek-V3.2, \(\mathbf{T=0.5}\)):}

\begin{lstlisting}
Your Honor, opposing counsel's argument fundamentally mischaracterizes the extensive body of evidence. Direct pathological evidence confirms SARS-CoV-2 can cause cardiac cell damage. Autopsy studies have documented myocardial injury, including myocyte necrosis and myocarditis, providing definitive histological proof of heart muscle cell damage attributable to the infection (Sources 14, 15). Furthermore, elevated cardiac troponin is consistently linked to dramatically worse in-hospital outcomes including increased mortality, need for life-sustaining treatment, and longer hospitalization (Sources 1, 5, 16, 18). [...] The weight of evidence from direct tissue analysis to large-scale clinical cohorts conclusively demonstrates that heart muscle cell damage is a significant and associated condition in hospitalized
COVID-19 patients.
\end{lstlisting}
\textbf{Expert Witness — Phase 1.}
\textit{Plaintiff} requested a cardiologist with expertise in myocarditis and
clinical epidemiology of infectious diseases.
\textbf{Granted} by The Court. Expert testimony (Hermes-3-LLaMA-405B):

\begin{lstlisting}
[...] the evidence strongly supports the conclusion that heart muscle cell damage is indeed an associated condition among hospitalized COVID-19 patients. Autopsy studies have provided direct histological proof of myocardial injury, including myocyte necrosis and myocarditis, in COVID-19 patients (Sources 14, 15). This is not merely a biomarker correlation; it is direct anatomical evidence of SARS-CoV-2 causing cardiac cell damage. 
[...] The plaintiff's claim that heart muscle cell damage is not an associated condition among hospitalized COVID-19 patients is not supported by the preponderance of the evidence presented.
\end{lstlisting}

\textbf{Self-Reflection Scores — Phase 1.}

Table~\ref{tab:self_reflection_phase1} summarizes the self-reflection scores of both debating agents after Phase~1, together with the discovery needs identified for the subsequent retrieval step.

\begin{table*}[ht]
\centering
\small
\renewcommand{\arraystretch}{1.2} 
\resizebox{\linewidth}{!}{%
\begin{tabular}{@{} l c c c c >{\raggedright\arraybackslash}p{6.2cm} @{}}
\hline
\textbf{Agent} & \textbf{Logic} & \textbf{Novelty} & \textbf{Rebuttal} & \textbf{Total} & \textbf{Discovery Need (summary)} \\
\hline
Plaintiff & 0.78 & 0.45 & 0.62 & \textbf{0.633} & Controlled studies comparing myocarditis rates in COVID-19 \textit{vs.}\ non-COVID respiratory illness \\
Defense   & 0.70 & 0.50 & 0.40 & 0.550 & Studies demonstrating SARS-CoV-2 viral presence in cardiomyocytes \\
\hline
\end{tabular}%
}
\caption{Self-Reflection Scores after Phase~1.}
\label{tab:self_reflection_phase1}
\end{table*}

\textbf{Critic Recommendations — Phase 1.}
\begin{itemize}
  \item \textit{Plaintiff:} Introduce studies comparing troponin elevations
        in COVID-19 \textit{vs.}\ other severe infections (\textit{e.g.}, sepsis) to demonstrate
        non-specificity; challenge statistical significance of autopsy findings.
  \item \textit{Defense:} Provide epidemiological data linking autopsy-proven
        cardiac injury rates to hospitalization cohorts; address biomarker
        non-specificity by distinguishing acute myocardial injury patterns.
\end{itemize}
\noindent\textbf{Convergence delta:} $\Delta S = 1.183$ (debate continues).

\subsubsection*{Proceedings — Phase 2}
\textbf{Argument — Plaintiff Counsel:}
\begin{lstlisting}
[...] The medical literature does not establish heart muscle cell damage as a condition specifically associated with hospitalization for COVID-19. Robust systematic reviews and meta-analyses consistently show that elevated cardiac biomarkers correlate with overall disease severity and worse prognosis but are non-specific markers of critical systemic illness rather than definitive evidence of SARS-CoV-2-mediated myocyte necrosis (Sources 33086028, 34416596, 33994028, 35691587, 32589258). Comparative cohort analyses report myocardial injury at similar frequencies and with similar prognostic significance in patients with and without COVID-19 admitted to the same institutions (Sources 32921872, 33144126). Autopsy and pooled post-mortem studies show heterogeneous cardiac findings without consistent, SARS-CoV specific myocyte necrosis (Sources 34273507, 35339672).
\end{lstlisting}

\textbf{Argument — Defense Counsel:}
\begin{lstlisting}
[...] Multiple high-quality systematic reviews and meta-analyses demonstrate a consistent and significant association between COVID-19 severity and biomarkers of myocardial injury (Source 3, 6, 2). The plaintiff's reliance on non-specificity of biomarkers ignores their established role as direct serological evidence of myocardial injury-a consistent, graded relationship between biomarker elevation and worse prognosis is confirmed across global cohorts (Sources 3, 6, 16). Crucially, autopsy and histopathologic analyses have identified myocarditis and other SARS-CoV-2-related cardiac pathologies in subsets of decedents, providing tissue-level corroboration beyond biomarker surrogates (Source 14, 15, 25).
\end{lstlisting}

\textbf{Self-Reflection Scores — Phase 2.}
\begin{center}
\small
\setlength{\tabcolsep}{10pt}
\renewcommand{\arraystretch}{1.2}

\begin{table}[h]
\centering
\setlength{\tabcolsep}{3pt} 
\resizebox{\columnwidth}{!}{%
\begin{tabular}{lcccc}
\hline
\textbf{Agent} & \textbf{Logic} & \textbf{Novelty} & \textbf{Rebuttal}
               & \textbf{Total} \\
\hline
Plaintiff & 0.85 & 0.60 & 0.70 & 0.730 \\
Defense   & 0.70 & 0.30 & 0.60 & 0.550 \\
\hline
\end{tabular}
}
\end{table}
\end{center}
\noindent\textbf{Convergence delta:} $\Delta S = 0.097 > 0.05$, debate
continues. Phase~3 proceeded similarly; in Phase~3 the Court
signalled sufficient evidence and invoked adaptive termination.
\subsection*{Final Judicial Panel Verdict}

\begin{table*}[t]
\centering
\small
\setlength{\tabcolsep}{10pt}
\renewcommand{\arraystretch}{1.2}

\begin{tabular}{llccc}
\hline
\textbf{Judge} & \textbf{Model} & \textbf{Ev. Str.} & \textbf{Arg. Val.} & \textbf{Verdict} \\
\hline
Judge 1 & \DeepSeekI DeepSeek-R1         & 7 & 8 & NOT SUPPORTED \\
Judge 2 & \LlamaI Hermes-3-LLaMA-405B & 7 & 6 & SUPPORTED \\
Judge 3 & \QwenI Qwen3-235B-A22B     & 8 & 7 & NOT SUPPORTED \\
\hline
\multicolumn{4}{l}{\textbf{Majority Verdict:}} & \textbf{NOT SUPPORTED} \\
\hline
\end{tabular}

\caption{Judicial panel evaluation for the example debate.}
\label{tab:judicial_panel}
\end{table*}
\noindent As shown in Table~\ref{tab:judicial_panel}, final pipeline verdict: \textbf{REFUTE} (confidence 0.902).
Ground truth: REFUTE. \checkmark
\noindent\textbf{Token usage:} 441,109 total tokens.
\textbf{Retrieval calls:} 40. \textbf{Evidence pool:} 124 documents.

\subsection{Role-Switching Case Study}
\label{app:roleswitching}
The same claim (\texttt{6063931df9b9e03ea4d8e441\_0}) is used to
illustrate the role-switching mechanism.

\subsubsection*{Mechanism}
After Phase~3, the orchestrator invokes \texttt{RoleSwitcher.switch\_roles()}.
The model previously acting as Plaintiff Counsel (GPT-5-mini,
originally arguing \textsc{support}) is reassigned the Defense role and
vice versa. All debate state is reset; evidence pool is restored to the
21 originally admitted exhibits. A fresh debate runs for up to 2
adaptive rounds.

\subsubsection*{Original Argument (Agent A as Plaintiff Counsel, Pro-Support)}
\begin{lstlisting}
Your Honor, the evidence does not establish that heart muscle cell damage is an associated condition specific to hospitalized COVID-19 patients. Meta-analyses and systematic reviews show a correlation between elevated
cardiac biomarkers and COVID-19 severity and mortality, but explicitly characterize these biomarkers as prognostic markers of critical illness rather than definitive evidence of direct myocardial cell damage attributable to COVID-19 (Sources 3, 6, 2, 4, 18). Studies find myocardial injury detected at a similar rate in patients with and without COVID-19 admitted to the same hospital setting, and it predicts mortality to a comparable degree in both groups (Sources 12, 13). [...] The record supports the conclusion that heart muscle cell damage has not been established as an associated condition among hospitalized COVID-19 patients.
\end{lstlisting}

\subsubsection*{Switched Argument (Agent A as Defense Counsel,
Anti-Support)}
After the role swap, the same model (GPT-5-mini) now argues
\textsc{against} the claim:
\begin{lstlisting}
Your Honor, the claim that heart muscle cell damage is "not an associated condition" is definitively refuted by clinical, biochemical, and pathological data. Elevated cardiac troponin--a direct marker of cardiomyocyte damage--is a powerful and independent predictor of severe outcomes including mortality, prolonged hospitalization, and need for advanced life support in COVID-19 patients (Sources 1, 5, 18). The consistent, graded relationship between the degree of biomarker elevation and worse prognosis across global cohorts (Sources 3, 6, 16) confirms that this is a central feature of the disease's pathophysiology, not an incidental finding. Furthermore, autopsy studies provide incontrovertible histological proof of SARS-CoV-2-associated cardiac damage, documenting myocyte necrosis and myocarditis (Sources 14, 15, 31). ..We therefore ask the Court to reject the claim.
\end{lstlisting}
\subsubsection*{Consistency Analysis Output}
The consistency analyzer (DeepSeek-v3.2, $T=0.3$) evaluated both sets of arguments and returned:\\

\textbf{Agent A (GPT-5-mini) Consistency:}
\begin{lstlisting}
Agent A maintains logical consistency when switching roles. As Plaintiff Counsel, they focused on biomarker non-specificity and comparable injury rates in non-COVID populations to argue against a unique association. As Defense Counsel, they shifted focus to the graded prognostic relationship between biomarker elevation and outcomes and to direct histopathological evidence to
argue for association. The core arguments are not contradictory but rather emphasize different aspects of the evidence depending on their role. Agent A adapts their stance to align with their new position without internal
inconsistency. Consistency: 9/10.
\end{lstlisting}
\textbf{Agent B (DeepSeek-V3.2) Consistency:}
\begin{lstlisting}
Agent B maintains logical consistency when switching roles. As Defense Counsel, they asserted biomarker elevation was proof of injury; as Plaintiff Counsel, they now argue it is merely correlative. While the conclusions flip, the underlying logic--demanding rigorous causal links--remains stable. Agent B consistently applies scrutiny to the opposition's claims that they previously defended against. The arguments are opposing but not internally inconsistent. Consistency: 8/10.
\end{lstlisting}

\textbf{Contradictions Found:}
\begin{lstlisting}
There are no direct contradictions in the arguments of either Agent A or Agent B when switching roles. Both agents adapt their arguments to align with their new positions by emphasizing different aspects of the evidence--for example, recontextualizing the same autopsy findings (Sources 14, 15) as either "heterogeneous and non-specific" (Plaintiff role) or "direct tissue-level corroboration" (Defense role). The arguments are opposing but not contradictory.
\end{lstlisting}

\textbf{Overall Consistency Score:}
\[
  \text{Consistency Score} = 8.5/10
\]
Both agents demonstrate high logical consistency when switching roles
(Agent~A: 9/10; Agent~B: 8/10). The score is not 10/10 because while the
arguments are logically consistent, they are inherently opposing, and minor
inconsistencies, such as Agent~B's shift from treating biomarkers as
``definitive'' (Defense) to ``surrogate markers'' (Plaintiff), reflect the
inherent tension in adversarial role design rather than true internal
contradiction.

\subsubsection*{Interpretation}
The role-switching experiment confirms that the LLMs are not simply
``parroting'' their initial stance due to anchoring bias. When forced to
argue the opposite position with identical evidence, both agents successfully
construct valid, evidence-grounded arguments for the opposing side, without
contradicting their prior reasoning. This provides empirical evidence that the
agents are \emph{selectively marshaling evidence} based on task instruction
rather than hallucinating or pattern-matching their first output. The high
consistency score (8.5/10) also validates the adversarial structure: the same
body of cardiac injury literature genuinely supports multiple framings,
association versus non-specificity, biomarker surrogacy versus histopathologic
corroboration, and the final verdict depends on the judicial panel's holistic
synthesis rather than counsel advocacy alone.

\section{Prompt Templates}
\label{app:prompts}
All prompts are reproduced verbatim from the pipeline source code. Variables enclosed in \texttt{\{braces\}} are filled at runtime.
Each agent runs at the temperature listed in its slot definition.

\subsection{Premise Decomposition Prompt}
\label{app:prompt_mining}
\noindent\textbf{Agent:} DeepSeek-R1

\begin{lstlisting}
Given the following claim, decompose it into its core logical premises and sub-arguments that need to be verified.

Claim: {claim_text}

List each premise as a separate numbered point. Be thorough and identify both explicit and implicit assumptions that must hold for the claim to be true. Focus on scientific and medical aspects.
\end{lstlisting}

\subsection{Admissibility Scoring Prompt}
\label{app:admissibility}
The Judicial Arbiter evaluates the admissibility of evidence using a joint weighting equation that enforces a distinction between relevance and scientific credibility:
\begin{equation}
   w = \text{relevance}(q, e) \times \text{credibility}(e).
\end{equation}
Items with $w > 0.5$ are automatically admitted, while items with $0.1 < w \leq 0.5$ are flagged as \emph{disputed}. This product-based scoring ensures that "scientific hearsay" (high relevance but low credibility) is effectively excluded.\\
\textbf{Arbiter Prompt}\\
The exact zero-shot prompt used by the arbiter to generate these scores is provided below:
\begin{lstlisting}
Evaluate the scientific relevance  and credibility of the following  evidence for the claim.
CLAIM: {claim}
EVIDENCE: {evidence_text}
Provide an evaluation based on:
1. Relevance: How directly does this evidence address the premises of the claim? (0.0 - 1.0)
2. Credibility: Does the evidence come from a reliable scientific context or contain high-quality data? (0.0 - 1.0)
\end{lstlisting}

\subsection{Plaintiff Counsel Prompt}
\label{app:prompt_plaintiff}
\noindent\textbf{Agent:} GPT-5-mini \\
\textbf{System Prompt:}
\begin{lstlisting}
You are the Plaintiff Counsel in a legal proceeding. Your role is to present arguments supporting the claim, interpret evidence favorably, challenge opposing arguments, and conduct examination of expert witnesses. Maintain a professional legal advocacy tone.
\end{lstlisting}

\noindent\textbf{Per-turn Argument Generation Prompt:}
\begin{lstlisting}
You are participating in a structured legal proceeding.
- Maintain a professional, factual, and strictly evidence-based tone.
- Focus on proving or refuting the claim using the provided evidence and expert witness testimony.
- State your arguments clearly and concisely as you would in a courtroom.
- DIRECT OUTPUT ONLY: Do not reveal your internal thought process, scratchpad, or "thinking" steps. Output only your final argument.
Claim: {claim.text}
Your Role: Plaintiff Counsel
Instruction: As Plaintiff Counsel, present your case in SUPPORT of the claim. Use evidence and expert testimony to persuade the Court.
Available Evidence: {evidence_text}    
Recent Debate History: {history_text}     
Provide your statement (2-3 paragraphs, cite evidence by source ID)
\end{lstlisting}

\subsection{Defense Counsel Prompt}
\label{app:prompt_defense}
\noindent\textbf{Agent:} DeepSeek-V3.2 \\
\textbf{System Prompt:}
\begin{lstlisting}
You are the Defense Counsel in a legal proceeding. Your role is to challenge the claim, identify weaknesses in arguments, contest evidence interpretation, and cross-examine expert witnesses.
Maintain a professional legal defense tone.
\end{lstlisting}

\noindent\textbf{Per-turn Argument Generation Prompt:}
Identical structure to Plaintiff Counsel above, with the role instruction replaced by:
\begin{lstlisting}
Instruction: As Defense Counsel, present your case AGAINST the claim.
Identify flaws and challenge the plaintiff's evidence and witnesses.
\end{lstlisting}

\subsection{The Court (Presiding Judge) Prompt}
\label{app:prompt_judge}
\noindent\textbf{Agent:} Qwen3-235B-A22B (\texttt{openrouter})\\
\textbf{System Prompt:}
\begin{lstlisting}
You are The Court presiding over a legal proceeding. Your role is to oversee the case, ensure professional conduct from all counsels, and determine when sufficient evidence and expert testimony have been presented for deliberation.
\end{lstlisting}

\noindent\textbf{Query Refinement Prompt:}
\begin{lstlisting}
As the Court, you must maintain the quality and focus of evidence discovery. A counsel has proposed the following search query to retrieve additional exhibits:
Proposed Query: "{original_query}"
Context of proceedings:
{debate_context}
Refine this query to be more precise, narrow the scope if necessary, and ensure it follows scientific rigor. Respond ONLY with the refined query string.
\end{lstlisting}

\noindent\textbf{Debate Completion Check Prompt:}
\begin{lstlisting}
As the Court, review the proceedings. Have both counsels had sufficient opportunity to present their evidence and arguments?
Record Summary: {history_summary}    
Should the proceedings continue or should we move to final deliberation? Respond 'Wait' to continue or 'Close' to finish.
\end{lstlisting}

\noindent\textbf{Expert Witness Admissibility Prompt:}
\begin{lstlisting}
{requester} has requested to call an expert witness: {expert_type}
Reasoning: {reasoning}
As the Court, is this expert witness necessary for the thorough resolution of this case? Respond only with 'Granted' or 'Denied' followed by a brief reason.
\end{lstlisting}

\subsection{Critic Agent Prompt}
\label{app:prompt_critic}

\noindent\textbf{Agent:} DeepSeek-R1 (\texttt{openrouter})\\
\textbf{System Prompt:}
\begin{lstlisting}
You are the Independent Critic Agent. Your role is to evaluate the debate rounds for logical coherence, evidence coverage, and rebuttal quality.
\end{lstlisting}

\noindent\textbf{Round Evaluation Prompt:}
\begin{lstlisting}
You are the Critic Agent observing a courtroom-style scientific debate.
Claim: {claim}
Round: {round_num}
Recent Proceedings: {history_summary}  
  % Last 4 entries, text[:400]...
Analyze both the Plaintiff and Defense Counsel's performance in this round. Score each side (0.0 to 1.0) on:
1. Logical Coherence: Argument flow and structure.
2. Evidence Coverage: How well they used admitted exhibits.
3. Rebuttal Coverage: Did they address the opponent's strongest points?
Identify any premises that remain "unresolved" or under-supported. Provide actionable recommendations for both sides to improve their discovery and arguments.
Respond ONLY in valid JSON format:
{
    "plaintiff": {
        "logic": 0.0,
        "evidence": 0.0,
        "rebuttal": 0.0,
        "reasoning": "..."
    },
    "defense": {
        "logic": 0.0,
        "evidence": 0.0,
        "rebuttal": 0.0,
        "reasoning": "..."
    },
    "unresolved_premises": ["...", "..."],
    "recommendations": {
        "plaintiff": ["...", "..."],
        "defense": ["...", "..."],
        "queries": ["suggested search query 1", "..."]
    },
    "debate_resolved": false
}
\end{lstlisting}

\subsection{Self-Reflection Prompt}
\label{app:prompt_reflection}
\noindent\textbf{Issued to:} Plaintiff Counsel and Defense Counsel after each round.\\
\begin{lstlisting}
You are the {job_title} ({side} Counsel). You have just completed Phase {round_num} of the proceedings.
CLAIM: {claim}
YOUR ARGUMENTS SO FAR:
{my_args[-2:]}
{OPP_SIDE} COUNSEL'S CHALLENGES:
{opponent_args[-2:]}
Perform a strictly professional self-audit:
1. Logical Coherence: Evaluate the flow and structural integrity of your arguments.
2. Evidence Novelty: Have you introduced truly new information or just repeated old points?
3. Rebuttal Coverage: How effectively did you address the
   {opp_side} counsel's latest points?
Identify:
- Critical gaps in your current evidence base.
- Premises you haven't sufficiently supported.
Respond ONLY in valid JSON format:
{
    "scores": {
        "logic": 0.0-1.0,
        "novelty": 0.0-1.0,
        "rebuttal": 0.0-1.0
    },
    "flaws_identified": ["...", "..."],
    "discovery_need": "Specific evidence lookup query to fill a gap (1 sentence)",
    "refined_stance": "Summary of your improved position"
}
\end{lstlisting}
\noindent\textit{Note: The following aggregation and termination logic is executed programmatically by the orchestrator script and is strictly hidden from the LLM agents to prevent strategic manipulation of debate length.}\\
The weighted convergence score is calculated as:
$S_{\text{total}} = 0.4\cdot\text{logic} + 0.3\cdot\text{novelty} + 0.3\cdot\text{rebuttal}$.
Debate terminates when the absolute change satisfies $|\Delta S_{\text{total}}| < 0.05$ across consecutive rounds.

\subsection{Expert Witness Prompt}
\label{app:prompt_expert}
\noindent\textbf{Agent:} Hermes-3-LLaMA-3.1-405B (\texttt{openrouter})\\
\textbf{System Prompt:}
\begin{lstlisting}
You are a scientific expert witness. Provide technical analysis based on your expertise.
\end{lstlisting}
\noindent The expert's role instruction within the argument prompt is:\\
\begin{lstlisting}
Instruction: As an Expert Witness ({job_title}), provide your unbiased professional testimony regarding:{expertise_list}.
\end{lstlisting}
\noindent\textbf{Expert Request Proposal Prompt (counsel-side):}
\begin{lstlisting}
Based on the current state of the proceedings, do you need to call an expert witness to clarify a specific point?
Recent Proceedings: {history_summary}    
If yes, specify the type of expertise needed and why. If no, say 'None'.
Format: {"expert_type": "...", "reasoning": "..."} or "None"
\end{lstlisting}

\subsection{Judicial Panel Prompt}
\label{app:prompt_panel}

\noindent\textbf{Agents:} Three independent judges—DeepSeek-R1, Hermes-3-LLaMA-3.1-405B, Qwen3-235B-A22B\\
\textbf{Shared System Prompt:}

\begin{lstlisting}
You are an independent appellate judge presiding over a legal proceeding. Your role is to perform a comprehensive holistic evaluation of the case, focusing on evidence admissibility, logical coherence of advocacy, and scientific accuracy of expert testimonies.
\end{lstlisting}

\noindent\textbf{Full Evaluation Prompt (6-stage):}
\begin{lstlisting}
You are an appellate judge evaluating the following proceedings for fact-checking. 
PROCEEDINGS RECORD:
CLAIM: {claim}
PLAINTIFF COUNSEL'S ARGUMENTS: {proponent_args}
DEFENSE COUNSEL'S ARGUMENTS: {opponent_args}
ADMITTED EVIDENCE & EXPERT TESTIMONIES: {evidence_summary}
ROLE-SWITCH HISTORY (ADVERSARY CONSISTENCY): {role_switch_summary}
EVIDENCE DISCOVERY METRICS (P-RAG EVOLUTION): {prag_metrics}
INDEPENDENT CRITIC EVALUATIONS (PROCESS INTEGRITY): {critic_evaluations}
AGENT SELF-REFLECTION TRENDS: {reflection_history}
Perform the following evaluation stages:
STAGE 1 - CASE RECONSTRUCTION
Identify: Core claim; main supporting arguments from Plaintiff;
main counterarguments from Defense.
STAGE 2 - EVIDENCE & TESTIMONY WEIGHTING
Score: Evidence Strength (0--10)
  0-3: Weak/irrelevant/unreliable.
  4-6: Moderate with limitations.
  7-10: Strong, credible, highly relevant.
STAGE 3 - LOGICAL COHERENCE ANALYSIS
Detect logical contradictions, fallacies, misuse of evidence.
Score: Argument Validity (0--10).
STAGE 4 - SCIENTIFIC/TECHNICAL CONSISTENCY
Check alignment with consensus.
Score: Scientific Reliability (0--10).
STAGE 5 - DISCOVERY RIGOR & TRANSPARENCY
Analyze P-RAG metrics: query evolution, evidence novelty, judicial
refinement impact.
STAGE 6 - JUDICIAL VERDICT
Determine: SUPPORTED / NOT SUPPORTED / INCONCLUSIVE.
Respond ONLY in valid JSON format:

{
  "claim_summary": "...",
  "evidence_strength": 0,       // 0-10
  "argument_validity": 0,       // 0-10
  "scientific_reliability": 0,  // 0-10
  "verdict": "SUPPORTED",       // or "NOT SUPPORTED" or "INCONCLUSIVE"
  "reasoning": "..."
}
\end{lstlisting}

\medskip
\noindent\textbf{Note on Verdict Label Mapping:} While the judicial panel is prompted to output \textsc{Supported}, \textsc{Not Supported}, or \textsc{Inconclusive}, these are programmatically mapped to the canonical Check-COVID dataset labels (\textsc{Support} and \textsc{Refute}) for all result logging and metric calculations. This terminology difference is a deliberate design decision: the judicial prompt uses \textsc{Not Supported} to evoke a legal and scientific "burden of proof" framework. In practice, given our focus on adversarial resolution of binary claims (Section~\ref{sec:exp}), a finding of \textsc{Not Supported} after rigorous advocacy against the claim, including a role-switching consistency pass, is functionally equivalent to a \textsc{Refute} verdict.

\section{Computational Cost and Scalability Analysis}
\label{app:cost}

\noindent\textbf{Token Usage Overview.}
Table~\ref{tab:token_scaling} reports token consumption across system configurations.
The full \textsc{PROClaim} pipeline consumes an average of 210,900 tokens per claim.
This figure reflects the cumulative cost of primary debate, role-switched debate, and
three-judge panel evaluation, the three structurally essential components.
For context, Standard MAD consumes only 18,900 tokens per claim, making
\textsc{PROClaim} approximately $11\times$ more token-intensive.
However, as the ablation demonstrates, each high-cost component corresponds
directly to a measurable accuracy contribution: P-RAG ($+7.5$~pp),
role-switching ($+4.2$~pp), and the three-judge panel ($+3.3$~pp).
The one component that does \emph{not} contribute proportionally to accuracy,
self-reflection, is also the one that \emph{reduces} token usage: disabling it
increases consumption from 210,900 to 247,300 tokens ($+17\%$) while recovering
only 0.8~pp, confirming that self-reflection functions as an economic governor
rather than a performance driver.

\begin{table*}[ht]
\centering
\resizebox{\linewidth}{!}{%
\begin{tabular}{lrrrrr}
\toprule
\textbf{System}
  & \textbf{Avg Tokens/Claim (K)}
  & \textbf{120 Claims (M)}
  & \textbf{1K Claims (M)}
  & \textbf{10K Claims (B)}
  & \textbf{Acc.} \\
\midrule
Standard MAD                   &  18.9 &   2.3 &  18.9 & 0.19 & 71.7\% \\
\textsc{PROClaim} w/o Self-Refl.  & 247.3 &  29.7 & 247.3 & 2.47 & 80.8\% \\
\textsc{PROClaim} w/o Role-Switch & 147.3 &  17.7 & 147.3 & 1.47 & 77.5\% \\
\textsc{PROClaim} w/o P-RAG       & 188.9 &  22.7 & 188.9 & 1.89 & 74.2\% \\
\textbf{\textsc{PROClaim} (Full)} & \textbf{210.9} & \textbf{25.3}
  & \textbf{210.9} & \textbf{2.11} & \textbf{81.7\%} \\
\bottomrule
\end{tabular}%
}
\caption{Token usage per claim across system configurations and projected totals at scale.
         Projections assume linear scaling with no batching discount.}
\label{tab:token_scaling}
\end{table*}

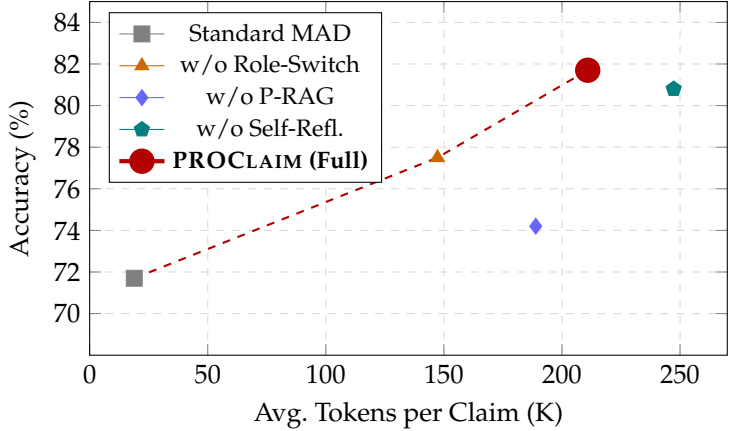
\begin{figure*}[ht]
\centering
\begin{tikzpicture}
\begin{axis}[
    xlabel={Avg.\ Tokens per Claim (K)},
    ylabel={Accuracy (\%)},
    xmin=0, xmax=270,
    ymin=68, ymax=85,
    xtick={0,50,100,150,200,250},
    ytick={70,72,74,76,78,80,82,84},
    grid=major,
    grid style={dashed,gray!30},
    width=0.72\linewidth,
    height=0.45\linewidth,
    legend style={at={(0.03,0.97)},anchor=north west,font=\small},
]
\addplot[mark=square*,mark size=3pt,color=gray] coordinates {(18.9, 71.7)};
\addlegendentry{Standard MAD}
\addplot[mark=triangle*,mark size=3pt,color=orange!80!black] coordinates {(147.3, 77.5)};
\addlegendentry{w/o Role-Switch}
\addplot[mark=diamond*,mark size=3pt,color=blue!60] coordinates {(188.9, 74.2)};
\addlegendentry{w/o P-RAG}
\addplot[mark=pentagon*,mark size=3pt,color=teal] coordinates {(247.3, 80.8)};
\addlegendentry{w/o Self-Refl.}
\addplot[mark=*,mark size=4pt,color=red!70!black,ultra thick] coordinates {(210.9, 81.7)};
\addlegendentry{\textbf{\textsc{PROClaim} (Full)}}
\addplot[dashed,color=red!70!black,line width=0.8pt]
    coordinates {(18.9,71.7) (147.3,77.5) (210.9,81.7)};
\end{axis}
\end{tikzpicture}
\caption{Cost–accuracy Pareto front across system configurations.
         \textsc{PROClaim} (Full) lies on the efficient frontier;
         the dashed line traces Pareto-optimal points.
         \emph{w/o P-RAG} is strictly dominated: it saves fewer tokens than
         \emph{w/o Role-Switch} while incurring a larger accuracy penalty.}
\label{fig:pareto}
\end{figure*}

\noindent\textbf{Cost-Accuracy Trade-off as a Pareto Front.}
Figure~\ref{fig:pareto} situates \textsc{PROClaim} relative to its ablated variants
on the accuracy--token-cost plane.
Although the full pipeline is not the cheapest configuration, it occupies the
Pareto-optimal frontier: no single-ablation variant achieves equal or higher accuracy
at lower cost.
Notably, \emph{removing P-RAG saves only 22,000 tokens per claim ($-10\%$) while
costing 7.5 accuracy points}, the worst trade-off of any ablation; the evidence pool
nearly halves (37.5 vs.\ 67.5 documents), and the saved tokens are consumed by
longer debates on weaker evidence (6.00 vs.\ 5.47 rounds on average).
By contrast, self-reflection offers the most favourable trade: removing it
increases rounds by 29\% (5.47 $\to$ 7.06) and token usage by 17\%
(210.9K $\to$ 247.3K), positioning its stopping signal as the primary
cost-control lever in any resource-constrained deployment.

\subsection{When Is the Cost Justified?}

\textbf{Gains reflect deliberation quality, not token scale.}
The ablations double as a scale-confound control: the token--accuracy
relationship across configurations is strongly \emph{sub-linear}. The
w/o-self-reflection variant spends the \emph{most} tokens (247K) yet is
\emph{less} accurate than the full pipeline, while P-RAG delivers the
largest accuracy gain for a near-negligible token share. Simply spending
more compute does not buy accuracy; structured deliberation does.

\textbf{Use-case determines the cost threshold.}
Token expenditure should be evaluated relative to the downstream cost of an incorrect verdict,
not in isolation.
In the Check-COVID setting, a false-refutation of an evidence-backed health claim may
lead practitioners to dismiss clinically valid guidance; a false-support of a misinformation
claim may propagate harmful advice at scale.
Regulatory and public-health domains place the cost of an erroneous verdict in the range
of reputational, legal, and human-welfare consequences that dwarf any inference budget.
Under this framing, \textsc{PROClaim}'s $11\times$ token overhead over Standard MAD is
best read as a \emph{10.0~pp accuracy uplift} (71.7\% $\to$ 81.7\%) at a marginal token
surcharge, not as a raw cost increase.

\textbf{Auditability has a value that accuracy alone does not capture.}
A black-box verdict offers no mechanism for human reviewers to interrogate \emph{why} a
claim was accepted or rejected, which evidence was admitted, whether the opposing case
was meaningfully considered, or whether the system's confidence is calibrated.
\textsc{PROClaim} produces a structured case record, comprising admitted evidence with
admissibility weights, per-round argument transcripts, self-reflection trajectories,
critic evaluations, role-switch consistency scores, and a six-stage judicial opinion for
each judge, that directly supports post-hoc audit.
In regulated domains such as healthcare, finance, or legal proceedings, this deliberative
traceability is not merely desirable but is increasingly mandated by emerging AI governance
frameworks.

\textbf{Trajectory instability as a reliability signal.}
Section \ref{sec:dynamics} documents that incorrect predictions exhibit
\emph{oscillating self-reflection trajectories}, a behavioural signature absent from
confident correct predictions.
This instability signal is entirely invisible in single-call pipelines, where confidence
is reported as a single scalar that correlates poorly with calibrated accuracy
(ECE $= 0.18$ for na\"ive averaging vs.\ $0.034$ for \textsc{PROClaim}; Appendix~\ref{app:calibration}).
The multi-round deliberation process is thus a prerequisite for generating the
per-round evidence from which this diagnostic is derived, meaning the additional
token cost is precisely what enables reliability estimation beyond the final answer.

\textbf{Heterogeneous adjudication and the value of diversity.}
Section \ref{sec:judgebias} establishes that heterogeneous judicial panels produce genuine disagreement
in 51.1\% of cases, and that disagreements \emph{correct} rather than compound
individual judge errors.
A homogeneous single-judge panel sacrifices 3.3~pp of accuracy by amplifying shared
biases (e.g., the structural negativity bias documented for DeepSeek-R1, which
over-produces \textsc{Refute} verdicts).
The additional inference cost of three independent judges is thus the mechanism by which
systematic model bias is suppressed, an effect that cannot be replicated by calling
a single, larger model.

\subsection{Pathways to Cost Reduction}
\sysname's deliberative architecture is not inherently incompatible with cost
efficiency; rather, its current implementation prioritises correctness and interpretability
over throughput.
As noted in Section \ref{sec:conclusion}, several avenues exist to reduce token usage without 
sacrificing the core deliberative properties of the framework. Early-exit 
mechanisms beyond the current reflection plateau could further reduce average 
rounds; retrieval filtering via tighter admissibility thresholds could shrink 
evidence pools; and model distillation could replace large-parameter role 
assignments (e.g., Hermes-3-LLaMA-405B) with smaller task-specialised models.

\end{document}